\documentclass[]{amap}

\usepackage[toc,page,header]{appendix}
\usepackage{minitoc}
\usepackage{solarized-light}
\usepackage{multirow}
\usepackage{graphicx}
\usepackage{siunitx,array}
\usepackage{makecell,array,booktabs}
\usepackage{framed}

\usepackage{amsmath}
\usepackage{amsfonts}
\usepackage{placeins}
\usepackage[colorinlistoftodos]{todonotes}
\usepackage{longtable}
\usepackage{hhline}
\usepackage{fancyvrb}
\usepackage{float}
\usepackage{fvextra}
\usepackage{CJKutf8}
\usepackage{multicol}
\usepackage{tablefootnote}
\usepackage{threeparttable}
\usepackage{mdframed}
\usepackage{subcaption}
\usepackage[usestackEOL]{stackengine}
\usepackage[numbers]{natbib}
\newcommand{\commentout}[1]{}
\usepackage{enumitem}
\setlist[itemize]{leftmargin=15pt}
\usepackage{siunitx,array}
\usepackage{tabularx}
\usepackage{adjustbox}
\usepackage{arydshln}
\usepackage{pifont}
\usepackage{bbding}
\usepackage{enumitem}
\definecolor{ampblue}{rgb}{0.82, 0.88, 0.94}
\usepackage[table]{xcolor} 
\usepackage{array} 
\usepackage[dvipsnames]{xcolor}
\usepackage{wrapfig}
\usepackage{todonotes}

\usepackage{xcolor}
\usepackage{hyperref}
\usepackage{cleveref}
\usepackage{tikz}
\usetikzlibrary{arrows.meta,positioning,fit,calc}

\hypersetup{
  colorlinks=true,
  linkcolor=blue!55!black,
  citecolor=blue!55!black,
  urlcolor=blue!55!black
}

\RequirePackage{xspace}
\makeatletter
\DeclareRobustCommand\onedot{\futurelet\@let@token\@onedot}
\def\@onedot{\ifx\@let@token.\else.\null\fi\xspace}

\makeatother

\definecolor{abot1}{HTML}{0185FE}
\definecolor{abot2}{HTML}{0185FE}
\definecolor{abot3}{HTML}{0185FE}
\definecolor{abot4}{HTML}{0185FE}
\definecolor{abot5}{HTML}{FB8C00}
\definecolor{abot6}{HTML}{FB8C00}
\definecolor{abot7}{HTML}{FB8C00}

\title{Behavior Foundations for Quadruped Robots: ABot-C0 Technical Report}

\author[*]{Xufeng~Zhao}
\author[*]{Fuzhi~Yang}
\author[*]{Jianhui~Chen}
\author{Li~Gao}
\author{Zhang~Meng}
\author{Jie~Gao}
\author{Yao~Zheng}
\author{Congyang~Zhao}
\author{Tianxiong~Lv}
\author{Menglin~Yang}
\author{Minqi~Gu}
\author{Yaru~Zhao}
\author{Wenyu~Liu}
\author{Honglin~Han}
\author{Shihui~Su}
\author{Zixiao~Tang}
\authorbreak
\author[\ddagger]{Liu~Liu}
\author{Mu~Xu}
\author[\ddagger]{Yang~Cai}
\author{Wenbin~Tang}

\affiliation{
\par\vspace{6pt}  
{\fontsize{11}{13}\selectfont \textbf{Amapbot Team, Amap, Alibaba Group}}
}

\renewcommand{\authorfont}{\fontsize{9}{11}\selectfont}
\renewcommand{\contributionfont}{\fontsize{9}{11}\selectfont}

\abstract{

The motion controller is one of the most fundamental modules in embodied intelligence systems. Driven by large-scale human motion-capture data and the motion-tracking paradigm, humanoid control has achieved remarkable progress in recent years. However, migrating this recipe to the quadrupedal setting is far less straightforward: animal motion data is scarcer and harder to capture at scale than human data, and cross-embodiment retargeting remains fragile. We present ABot-C0, a generalist motion-control system for quadruped robots that establishes three complementary behavior foundations: a scalable multi-source motion-data pipeline, robust policy learning across motion tracking, locomotion, and scene interaction, and a unified deployment stack for reliable real-world operation. Fundamentally, we construct a data pyramid through conditional video-generation synthesis, annotated motion capture, teleoperation, and human design, producing 16,074 physically feasible motion clips as the data foundation for diverse motion-learning demands. With large-scale motion data, a Flow-Matching generalist policy demonstrates, for the first time, a scaling law for quadruped motion tracking: performance improves consistently as training scales up, with zero-shot capability to track unseen motions. We then go a step further toward robust all-terrain locomotion by adopting a three-stage privileged-to-perceptive framework with temporal LiDAR memory and terrain-predictive supervision. Collectively, these components form a motion generalist that coordinates multi-policy execution, smooth behavior transitions, energy-efficient control, and safety mechanisms for real-world deployment. Extensive experiments on urban-terrain autonomous navigation and companion-style multimodal interaction demonstrate that quadruped robots can move beyond functional demos toward product-level behavioral intelligence.

}

\begin{document}
\maketitle
\makeatletter
\begingroup
\renewcommand{\@makefntext}[1]{\parindent=0pt\noindent #1}
\renewcommand{\thefootnote}{}
\footnotetext{\fontsize{7}{9}\selectfont
\begin{tabular}[t]{@{}l@{}}
$^*$Equal contribution.\\
$^\ddagger$Corresponding author: \{diana.ll, yangcai.cy\}@alibaba-inc.com
\end{tabular}}
\endgroup
\renewcommand{\thefootnote}{\arabic{footnote}}
\makeatother

\section{Introduction}

\begin{figure*}[t]
    \centering
    \includegraphics[width=\textwidth]{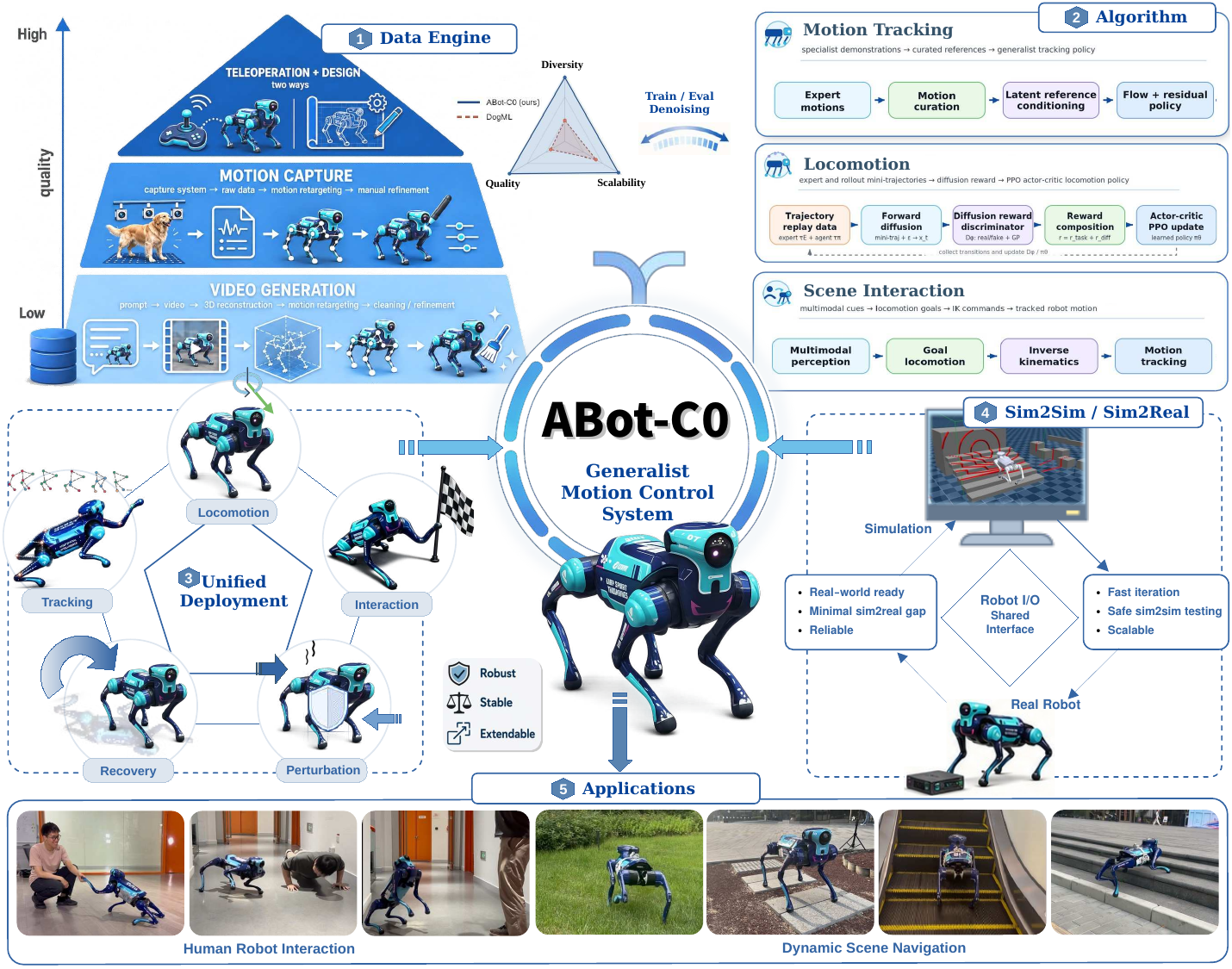}
    \caption{\textbf{Overview of ABot-C0.} ABot-C0 is organized as a generalist quadruped motion-control system built on a scalable data pyramid, a three-task learning suite for locomotion, motion tracking, and scene interaction, and a unified deployment layer that supports smooth policy switching, recovery, and perturbation robustness. Sim-to-sim verification precedes sim-to-real transfer through a shared robot interface, enabling extensible applications such as human-robot interaction and dynamic-scene navigation.}
    \label{fig:overview}
\end{figure*}

Motion control, the ability to translate high-level intent into physically feasible whole-body movements while maintaining stability, adaptability, and safety throughout physical interaction with the environment, is fundamental to realizing general-purpose robots in real-world applications.
  In recent years, reinforcement learning has become the dominant paradigm in this domain because of its flexibility and scalability \cite{peng2021amp,peng2022ASELargeScaleReusable,liao2025BeyondMimicMotionTracking}, and the rapid maturation of physics simulators~\cite{mujoco,isaacgym,isaaclab,genesis,newton} has further accelerated progress by enabling massively parallel policy optimization and narrowing the sim-to-real gap.
  Inspired by scaling laws in large language models, the community has begun to explore whether simultaneously scaling motion data, model capacity, and compute can yield generalist controllers with emergent capabilities. In the humanoid domain, this pursuit has given rise to a new class of \emph{Behavioral Foundation Models} (BFMs) for general whole-body control, with representative works including Sonic~\cite{sonic}, BFM-Zero~\cite{bfmzero}, and HoloMotion-1~\cite{holomotion}.

  A critical driver of these advances is large-scale motion data from multiple pipelines, e.g., MoCap systems~\cite{amass,bonesseed,phuma,twist2} and SMPL-based video reconstruction~\cite{videomimic}, that together supply massive diverse references. 
  Neither pipeline, however, transfers well to quadrupeds due to morphological mismatches and lack of diversity and behavioral intelligence in captured motions.
  In recent years, quadruped control has mainly focused on isolated sub-skills such as all-terrain locomotion~\cite{DreamWaQ,miki2022wild}, biomimetic gaits~\cite{peng2021amp,margolis2023walk}, and extreme agility~\cite{zhuang2023parkour,cheng2024extreme}. However, as the legged form closest to real-world deployment, quadruped robots demand far more than individual skills: we envision them evolving from merely \textit{``being able to walk''} to \textit{``being able to understand, express, and interact''}, a goal that demands a generalist motion-control system unifying behavioral foundations of motion generation, tracking, locomotion, scene interaction, and robust deployment into a single coherent stack, which leads to \textbf{ABot-C0} (see Figure~\ref{fig:overview}), our comprehensive effort toward building a Behavior Foundation Model (BFM) for quadruped robots. Our main contributions are summarized as follows:

  \begin{itemize}
    \item \textbf{Data Engine}: As the foundation of the system, we construct a scalable quadruped motion-data pyramid that combines motion capture, teleoperation, human-designed motions, and a controllable video-generation pipeline driven by text and image prompts. This pipeline produces high-quality, diverse, and scalable motion corpora with natural-language annotations, providing a data foundation for downstream motion learning tasks.
  
    \item \textbf{Quadruped General Motion Tracking}: Building on this data foundation, we present the first generalist motion-tracking controller for quadruped robots, trained on our large-scale motion dataset. Leveraging this controller, we empirically verify a scaling law for quadruped motion tracking with respect to the data volume. Furthermore, we introduce a novel Manifold-Calibrated Reference Conditioning (MCRC) method that enhances performance by conditioning the reference on a learned motion manifold, achieving over 90\% success rate with zero-shot generalization.
  
    \item \textbf{Versatile Locomotion}: Complementing reference-motion tracking, we build a progressive locomotion stack that first establishes robust velocity tracking, then extends it to biomimetic omnidirectional gait control through a diffusion-based stylistic prior and symmetric command augmentation, and finally transfers the controller to unstructured terrain using a three-stage privileged-to-perceptive framework with temporal LiDAR memory and terrain-predictive supervision.

    \item \textbf{Unified Multi-Task Deployment and Application Validation}: Finally, we integrate these capabilities into a unified quadruped control system that accepts multimodal commands, including language instructions, velocity commands, and spatial goals. The deployment stack coordinates smooth transitions across locomotion, motion tracking, and interaction policies while supporting recovery and perturbation robustness in changing environments. We validate the complete system through high-level applications, including all-terrain autonomous navigation and home-companion-style human-robot interaction, demonstrating its effectiveness, robustness, and extensibility.
  \end{itemize}

  \paragraph{Report organization} Section~\ref{sec:data_engine} presents the ABot-C0 data engine. Section~\ref{sec:tasks} describes the core motion capabilities and their methods, including motion tracking, locomotion, and scene interaction. Section~\ref{sec:system} introduces the unified deployment system. Section~\ref{sec:experiments} reports controlled evaluations, Section~\ref{sec:applications} demonstrates downstream applications, and Section~\ref{sec:conclusion} concludes with limitations and future directions.
\section{Data Engine}
\label{sec:data_engine}

High-fidelity motion tracking relies on a large and diverse reference motion library, yet existing quadruped motion-capture datasets are orders of magnitude smaller than their humanoid counterparts and limited in behavioral coverage. To address this bottleneck, our data engine builds a data pyramid (c.f. Figure~\ref{fig:overview} top left) from four complementary sources: \emph{Teleoperation}, \emph{Artist Design}, \emph{Motion Capture}, and \emph{Video-to-Motion Generation}. Teleoperation provides high-quality, robot-executable demonstrations for cold-start policy training, but is limited in scale and behavioral coverage. Artist-designed motions expand the library toward expressive and highly dynamic behaviors, although their physical feasibility is not guaranteed before downstream validation. Motion-capture data contributes natural, animal-like locomotion patterns, but offers limited control over trajectory, behavior composition, and collection scale. Video-based generation~\cite{liu2026unleashing} complements these sources as a scalable and controllable data backbone: text- and image-conditioned generation can grow the motion library with compute while covering behaviors that are difficult to collect, design, or retarget manually.
The remainder of this section focuses on the video-based generation pipeline and the subsequent multi-stage quality filtering that produces physically validated, language-annotated motion trajectories at scale.

\subsection{Video-to-Motion Generation}
\label{sec:video_gen}

The video-based branch turns natural-language and motion prior image prompts into deployable quadruped motion trajectories through three fully automated stages detailed below.

\subsubsection{Identity-Consistent Video Generation}
\label{sec:identity_consistent}

The key technical challenge is \textit{identity drift}: current video diffusion models deform the robot's body non-rigidly across frames, shifting its appearance and violating the rigid-body assumption that downstream 3D extraction depends on.
We fine-tune Wan2.2~\cite{wan2025} in a first-frame image-to-video (I2V) setup, where each clip is conditioned on a canonical reference image $I_{\mathrm{ref}}$ of the our legged robot. The I2V setting fixes the camera and background by construction, but does not prevent frame-to-frame identity shift.
We close this gap with an Identity Consistency Loss $\mathcal{L}_{\mathrm{IC}}$ layered on top of the standard flow-matching objective $\mathcal{L}_{\mathrm{FM}}$.

An appearance bank $\mathcal{B} = \{f_{\mathrm{ref}}^{(j)}\}_{j=1}^{N}$ is built via greedy coverage-set search over DINOv2~\cite{oquab2024dinov2} CLS embeddings of the training data (cosine threshold $\tau{=}0.8$).
At each training step, we decode the predicted clean latent through the frozen VAE, extract per-frame DINOv2 features $f_t$, and apply a nearest-reference hinge:
\begin{equation}
\mathcal{L}_{\mathrm{IC}} = \frac{1}{T}\sum_{t=1}^{T}\max\!\left(0,\; m_{\mathrm{id}} - \max_{j\in[N]}\cos(f_t, f_{\mathrm{ref}}^{(j)})\right).
\end{equation}
The final training objective is $\mathcal{L}_{\mathrm{total}} = \mathcal{L}_{\mathrm{FM}} + \lambda\,\mathcal{L}_{\mathrm{IC}}$, with $\mathcal{L}_{\mathrm{IC}}$ attached only to Wan2.2's low-noise expert where the predicted clean video is perceptually reliable enough for DINOv2 to score.

\subsubsection{From Video to 3D Motion Trajectory}
\label{sec:video_to_3d}

Each generated video is converted into a 3D motion trajectory $\{\mathbf{s}_t\}_{t=1}^{T}$ in the robot's configuration space. Although recovering 3D pose from monocular video is generally ill-posed, the I2V setup removes key ambiguities: camera intrinsics and extrinsics are known and fixed across all frames, and the frame-0 pose is exactly the URDF canonical standing state. This reduces trajectory recovery to a temporally constrained kinematic fitting problem over per-frame joint angles and root motion.

For each frame we predict 2D positions of $K$ predefined body landmarks using a ViTPose~\cite{xu2022vitpose} model fine-tuned on rendered robot images. We then solve for the per-frame state $\mathbf{s}_t = (\mathbf{p}_t, \boldsymbol{\phi}_t, \boldsymbol{\theta}_t)$, comprising the root position $\mathbf{p}_t \in \mathbb{R}^3$, root Euler angles $\boldsymbol{\phi}_t \in \mathbb{R}^3$, and 12 actuated joint angles $\boldsymbol{\theta}_t \in \mathbb{R}^{12}$, by minimizing a 2D reprojection error:
\begin{equation}
L_{\mathrm{reproj}}^{(t)} = \sum_{k} \| \Pi(\mathrm{FK}(\mathbf{s}_t)_k) - \mathbf{p}_{2D,k}^{(t)} \|_2^2,
\end{equation}
where $\Pi$ is the projection under the fixed camera and $\mathrm{FK}$ is the robot's URDF forward kinematics chain. We complement this with a temporal smoothness penalty and foot-contact constraints that anchor detected stance feet to the ground plane, preventing drift and sliding during contacts.

\subsubsection{Multi-Stage Quality Filtering}
\label{sec:quality_filter}

The trajectory extraction above does not guarantee success on every clip. We apply three sequential gates with increasing cost, so that cheap checks eliminate the bulk of failures before expensive simulation is invoked:
\begin{itemize}

    \item{\textbf{CLIP semantic gate.}}
    We re-render each recovered trajectory in MuJoCo under the original camera viewpoint and compute the mean CLIP~\cite{clip} image-embedding cosine similarity between sampled re-rendered and original frames.
    Trajectories below the similarity threshold are discarded (97.0\% pass rate).

    \item \textbf{Geometric gate.}
    Trajectories passing the semantic gate are further screened by reprojection error: per-clip mean $<20$\,px and per-clip max-frame $<100$\,px.
    This is the primary bottleneck in the pipeline (70.2\% pass rate), reflecting a deliberately conservative threshold---over-generating and discarding is safer than relaxing thresholds and propagating extraction artifacts into policy training.

    \item \textbf{Physical feasibility gate.}
    Even geometrically faithful trajectories may be physically infeasible. For each surviving trajectory, we train a per-motion specalist tracking policy using the controller described below and run a full-length rollout in simulation.
    Trajectories triggering any termination condition (fall, root divergence, velocity explosion) are discarded (97.6\% pass rate).

\end{itemize}

\subsection{Dataset Summary}
\label{sec:dataset}

Through the combination of all four data sources\cite{gao2026quadfm,liu2026unleashing}, we construct a motion library of \textbf{16,074}  physically validated reference trajectories totaling \textbf{22.43 hours}, each accompanied by language annotations, in Table~\ref{tab:dataset_breakdown}. Among them, the video-based generation pipeline alone contributes 7,488 clips (18.51 hours) spanning acrobatic maneuvers, expressive behaviors, and ground interactions. The full dataset serves as the data foundation for motion tracking (Section~\ref{sec:motion_track}) and locomotion (Section~\ref{sec:locomotion}) described in subsequent sections.

\begin{table*}[!ht]
\centering
\caption{ABot-C0 Dataset Statistics}
\resizebox{\textwidth}{!}{
\label{tab:dataset_breakdown}
\begin{tabular}{ccccccc}
\toprule
\multirow{2}{*}{\textbf{Motion Category}} & \multirow{2}{*}{\textbf{Data Source}} & \multirow{2}{*}{\textbf{Semantic Subset / Behavior Types}} & \multicolumn{2}{c}{\textbf{Ours}} & \multicolumn{2}{c}{\textbf{DogML}} \\
\cmidrule(lr){4-5} \cmidrule(lr){6-7}
 &  &  & \begin{minipage}{2cm}\centering\textbf{Clips} \end{minipage}& \begin{minipage}{2cm}\centering \textbf{Duration}\end{minipage} & \begin{minipage}{2cm} \centering \textbf{Clips} \end{minipage}&  \begin{minipage}{2cm}\centering\textbf{Duration}\end{minipage} \\
\midrule
\textbf{Locomotion}
 & Motion Capture & Basic Gaits (Walking, Trotting, etc.) & 7,998 & 10.02 h & 4,024$^*$ & 11.03 h \\
\midrule
\multirow{3}{*}{\begin{tabular}[c]{@{}l@{}}\textbf{Expressive} \\ \textbf{Behavior}\end{tabular}}
 & Teleoperation & Cold-Start Bootstrapping & 547 & 3.73 h & - & - \\
 & Artist Design & S-Tier Extreme Maneuvers & 41 & 0.19 h & - & - \\
 & Video Generation & Acrobatic, Expressive, Diverse Behaviors & \textbf{7,488} & \textbf{18.51 h} & - & - \\
\midrule
\textbf{Total} & \textbf{--} & \textbf{All Sources / Comprehensive} & \textbf{16,074} & \textbf{22.43 h} & \textbf{4,024$^*$} & \textbf{11.03 h} \\
\bottomrule
\multicolumn{7}{l}{\rule{0pt}{3ex}\footnotesize $^*$ \textit{Note:} DogML\cite{t2qrm} count includes redundant retargeted sequences. For a consistent comparison of behavioral diversity, we report the \textbf{4,024} unique motion events.}
\end{tabular}%
}
\vspace{-1em}
\end{table*}
\section{Core Motion Capabilities}
\label{sec:tasks}

\subsection{Motion Tracking}
\label{sec:motion_track}


Recent large-scale humanoid trackers and general motion tracking frameworks
have shown promising scalability across diverse motion libraries and skills~\cite{he2024omnih2o,ji2024exbody2,chen2025gmt,wang2025expertsgeneralist,sonic,he2025asap,liao2025BeyondMimicMotionTracking,zhang2025trackanymotions,cheng2024expressive,fu2024humanplus},
but whether this paradigm transfers to quadrupeds remains unclear. To address this gap, this work presents the first large-scale generalist
motion-tracking
controller for quadruped robots. In building this controller over thousands of
reference motions, we empirically verify a scaling law for quadruped motion
tracking: increasing the number of training motions systematically improves
unseen tracking performance and reduces the seen-unseen gap. Based on a deeper
analysis of the reference library, we further introduce Manifold-Calibrated
Reference Conditioning method (MCRC), which denoises physical violations and augments
the policy with a learned motion-manifold code and reconstruction uncertainty,
resulting in improved performance of the quadruped generalist controller. Overall, our
contributions in motion tracking are threefold:
\begin{itemize}
  \item To the best of our knowledge, we introduce the first generalist motion-tracking
  controller for quadruped robots, trained on a large-scale dataset.
  \item We verify the scaling law of quadruped motion tracking with respect to the data volume.
  \item We propose MCRC that improves generalist tracking performance by conditioning the policy on a learned motion manifold.
\end{itemize}

\begin{figure}[t]
\centering
\includegraphics[width=\linewidth]{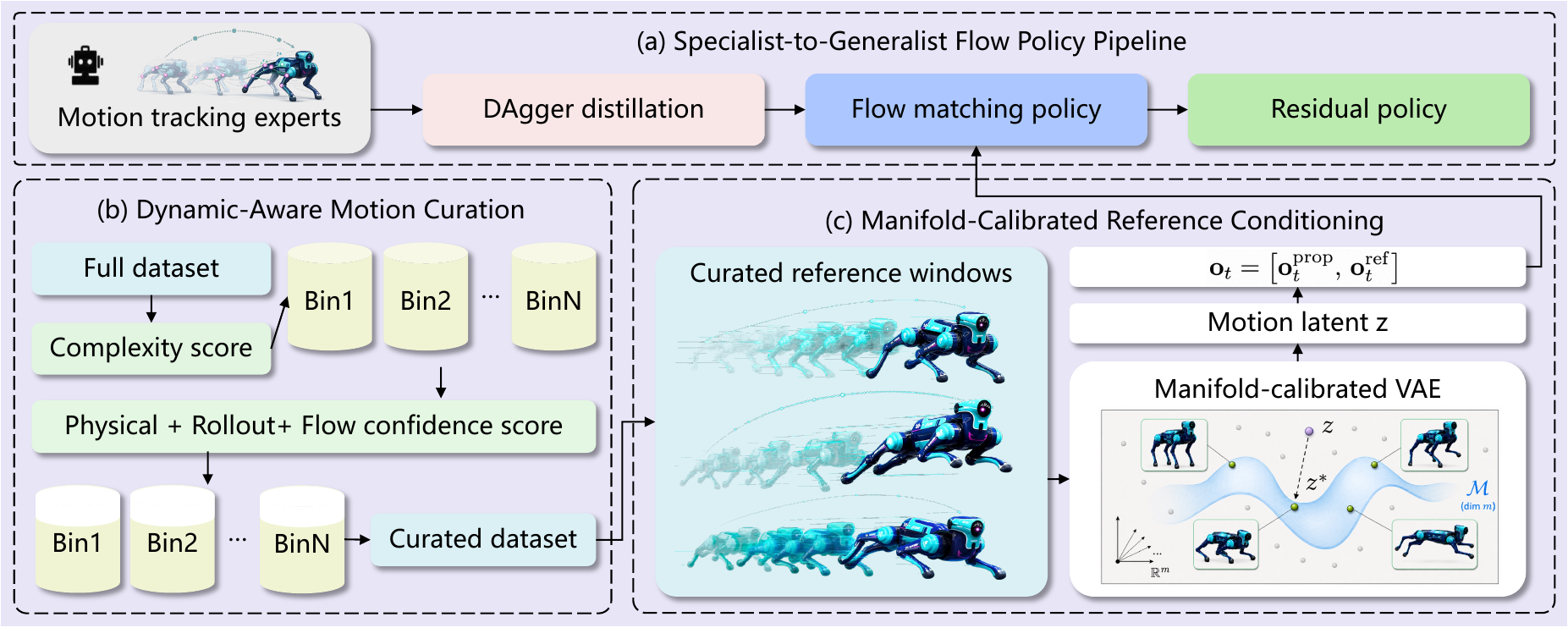}
\caption{Overview of the proposed quadruped general motion-tracking framework.
The framework consists of three main components: a specialist-to-generalist
flow policy pipeline with DAgger and residual RL, Dynamic-Aware Motion Curation
for selecting high-value reference motions, and Manifold-Calibrated Reference
Conditioning for augmenting the student policy with learned reference-manifold
signals.}
\label{fig:motion_tracking_method}
\end{figure}

\subsubsection{Methodology}
\label{sec:motion_methodology}

\paragraph{Specialist-to-Generalist Flow Policy Pipeline}
We adopt a specialist-to-generalist training strategy, which has been successfully applied to humanoid motion tracking~\cite{omnixtreme}, and adapt it to large-scale quadruped motion tracking (see Figure~\ref{fig:motion_tracking_method}).
Given a motion library
$\mathcal{M}=\{m_1,\ldots,m_M\}$, we first train one specialist policy
$\pi^k_{\mathrm{expert}}$ for each motion clip $m_k$ using PPO~\cite{ppo} in parallel simulation.
Each specialist receives the same single-frame tracking observation and is
optimized only for its assigned motion. This avoids the gradient interference
observed when one from-scratch multi-motion policy must simultaneously fit
reference clips with substantially different dynamics.
After sufficient training, the specialist set is then distilled into a unified policy $\pi_\theta$ using
DAgger~\cite{dagger}. The student rolls out in parallel environments, and the
visited states are relabeled by the corresponding per-motion specialists. This keeps supervision on the student-induced state distribution while keeping the specialist query interface fixed. The unified student can then introduce additional reference-side conditions without changing the expert policies used for DAgger supervision.

The student policy is parameterized as a conditional Flow-Matching (FM) model, because of its capability to represent multimodal expert-action distributions induced by diverse reference motions while retaining fast inference through a small number of ODE steps.
Let $\mathbf{a}_{\mathrm{expert}}$ be the specialist action for observation $\mathbf{o}$, and let $\epsilon\sim\mathcal{N}(0,I)$ be a base noise sample. Flow matching learns a time-dependent velocity field $v_\theta(\mathbf{a}_t,t,\mathbf{o})$ that transports noisy actions toward the specialist-action distribution along the linear path
\begin{equation}
  \mathbf{a}_t = (1-t)\mathbf{a}_{\mathrm{expert}} + t\epsilon,
  \quad t\sim\mathrm{Beta}(1.5,1.0),
\end{equation}
with loss
\begin{equation}
  \mathcal{L}_{\text{FM}}(\theta)
  =
  \mathbb{E}_{t,\epsilon,\mathbf{a}_{\text{expert}}}
  \left[
    \left\|
      v_\theta(\mathbf{a}_t,t,\mathbf{o})
      -
      \underbrace{(\epsilon-\mathbf{a}_{\text{expert}})}_{\text{target velocity}}
    \right\|^2
  \right].
\end{equation}
This objective turns action prediction into conditional denoising dynamics:
the policy starts from a sampled action noise and repeatedly follows the learned
velocity field toward the action manifold induced by the specialist ensemble.
At inference, actions are obtained by reverse Euler integration:
\begin{equation}
  \mathbf{a}_{t-1/D}
  =
  \mathbf{a}_t
  -
  \frac{1}{D}v_\theta(\mathbf{a}_t,t,\mathbf{o}),
  \quad
  t=1,\frac{D-1}{D},\ldots,\frac{1}{D}.
\end{equation}

We further train a residual RL policy on top of the frozen flow policy to
improve robustness and recover performance gaps that remain after distillation.
The residual actor receives the current tracking observation and the base flow
action, and produces a bounded correction:
\begin{equation}
  \mathbf{a}_{\mathrm{total}}
  =
  \mathbf{a}_{\mathrm{flow}}
  +
  s\cdot
  \mathrm{clip}(\Delta \mathbf{a}, \pm c),
\end{equation}
where $c=0.5$ and $s=0.2$, in practice, denote the residual clip threshold and correction scale.
The residual stage is intentionally lightweight: the base flow policy already solves
the tracking task well in simulation, so prolonged residual PPO training can
drift away from the high-fidelity distilled behavior.

\paragraph{Dynamic-Aware Motion Curation}
The motion library is heterogeneous in quality and closed-loop executability. To preserve behavioral coverage, we first stratify motions by a complexity measure
\begin{equation}
  c(m):=\mathrm{clip}(\mathbf{w}_{c}^{\top}\mathbf{u}^{\mathrm{cmp}}(m),0,1),
\end{equation}
where $\mathbf{u}^{\mathrm{cmp}}(m)$ contains normalized features over root motion, height variation, and joint motion range. Motions are then divided into equal-frequency bins according to $c(m)$.

Inside each bin, references are ranked by three complementary criteria. Physical feasibility penalizes normalized violations in joint dynamics, root motion, base tilt, foot sliding, and contact consistency:
\begin{equation}
  p(m)=\exp(-\mathbf{w}_{p}^{\top}\mathbf{v}^{\mathrm{phys}}(m)).
\end{equation}
Rollout executability combines the fixed-policy success rate $s(m)$ with normalized tracking error $\bar{e}(m)$:
\begin{equation}
  r(m) = s(m)\exp(-\bar{e}(m)).
\end{equation}
Flow confidence measures whether the distilled flow policy produces consistent actions under repeated sampling:
\begin{equation}
  f(m) = \exp(-\gamma_f\sigma_{\mathrm{flow}}(m)),
\end{equation}
where $\sigma_{\mathrm{flow}}(m)$ is the action variance across multiple flow samples for the same observation. Lower variance therefore corresponds to higher confidence. We call the resulting selector PRF-score selection, for Physical feasibility, Rollout executability, and Flow confidence. The bin-wise ranking score is

\begin{equation}
  S_{\mathrm{cur}}(m)
  =
  \lambda_p p(m) + \lambda_r r(m) + \lambda_f f(m),
  \qquad
  \lambda_p+\lambda_r+\lambda_f=1.
\end{equation}

In our implementation, we set $\lambda_p=0.45$, $\lambda_r=0.35$, and $\lambda_f=0.20$ for physical feasibility, rollout executability, and flow confidence, respectively. This preserves complexity coverage while preferring references that are physically plausible, executable, and not highly uncertain under the flow sampler. The same curated references also provide a cleaner distribution for learning the VAE reference manifold used in the next module. Section~\ref{sec:dynamic_aware_motion_curation_eval} evaluates the contribution of each sampling criterion and their combined PRF score under a fixed motion budget.

\paragraph{Manifold-Calibrated Reference Conditioning (MCRC)}
Motion curation improves the quality of the reference set, but it does not by
itself change the information available to the distilled student policy. In
particular, the reference-side input is still a frame-level tracking command and
therefore provides no explicit descriptor of the local reference segment's
location on the learned motion manifold. To provide this missing context, we learn a compact
reference-manifold representation with a reference-window VAE and use the
resulting time-indexed latent code as an additional student-side condition. We also retain
the VAE reconstruction uncertainty as an auxiliary reliability signal for
ablation analysis.

The VAE is trained on curated reference windows, thus its latent space is shaped
by motions that are physically plausible, executable, and representative of the
training distribution. For motion $m$, we denote by $\mathbf{x}_{m,t:t+H}$ the length-$H$ reference window starting at time $t$, formed by concatenating the normalized kinematic reference features used for tracking, including reference joint poses, root-relative body poses, and body velocities. The encoder parameterizes a Gaussian posterior
$q_\psi(\mathbf{z}\mid\mathbf{x}_{m,t:t+H})=\mathcal{N}(\boldsymbol{\mu}_{m,t},\mathrm{diag}(\boldsymbol{\sigma}^2_{m,t}))$,
while the decoder reconstructs the input window as $\hat{\mathbf{x}}_{m,t:t+H}$. The training objective combines reconstruction with a KL regularizer,
\begin{equation}
  \mathcal{L}_{\mathrm{VAE}}
  =
  \left\|\hat{\mathbf{x}}_{m,t:t+H}-\mathbf{x}_{m,t:t+H}\right\|_2^2
  +
  \beta\,D_{\mathrm{KL}}
  \left(
    q_\psi(\mathbf{z}\mid\mathbf{x}_{m,t:t+H})
    \,\|\,\mathcal{N}(0,I)
  \right),
\end{equation}
which regularizes the posterior toward a smooth latent space while preserving
the local kinematic structure of each reference window.

At each control step, the reference condition is computed from the lookahead
window that starts at the current reference frame. We use the posterior mean of
this window as the manifold coordinate:
\begin{equation}
  \mathbf{z}_t(m)
  =
  \boldsymbol{\mu}_{m,t}.
\end{equation}
Near the end of a clip, the final reference frame is repeated to form a
length-$H$ window. Thus $\mathbf{z}_t(m)$ describes the local future segment
$\mathbf{x}_{m,t:t+H}$. We also use the corresponding per-window reconstruction
error,
\begin{equation}
  e_{\mathrm{recon},t}(m)
  =
  \left\|
    \hat{\mathbf{x}}_{m,t:t+H} - \mathbf{x}_{m,t:t+H}
  \right\|_2^2,
\end{equation}
which provides a reliability signal for how well the current lookahead window is
explained by the learned manifold.

In our final student policy, we use the dynamic manifold coordinate as the
additional future-reference condition:
\begin{equation}
  \mathbf{o}^{\mathrm{student}}_t
  =
  \left[
    \mathbf{o}_t,\,
    \mathbf{z}_t(m)
  \right],
\end{equation}
The reconstruction signal $e_{\mathrm{recon},t}(m)$ is not part of the default
observation, but we evaluate it as an auxiliary condition, both alone and
together with $\mathbf{z}_t(m)$, in the ablation study in
Section~\ref{sec:experiments}.

\subsubsection{Implementation Details}
\label{sec:motion_training_details}

\noindent\textbf{Observation and Action Space.}
The base tracking interface uses a 69-dimensional observation vector
$\mathbf{o}_t=[\mathbf{o}^{\mathrm{prop}}_t,\mathbf{o}^{\mathrm{ref}}_t]$, which includes proprioceptive state and motion-reference observations. The proprioceptive block contains relative joint positions
$\mathbf{q}_t-\mathbf{q}_0\in\mathbb{R}^{12}$, joint velocities
$\dot{\mathbf{q}}_t\in\mathbb{R}^{12}$, base angular velocity
$\boldsymbol{\omega}_t\in\mathbb{R}^3$, and a 6D representation of the torso orientation difference between the robot and the reference frame. The reference block provides the current-frame tracking command, including reference joint positions, root-relative body positions, and body velocities extracted from the motion clip. This base observation is used by all specialist policies and by the expert query interface during DAgger.

For the Transformer-based flow policy, the 69-dimensional base
observation is parsed as a 24-dimensional reference-command block and a
45-dimensional proprioceptive block. The history stores only the
45-dimensional proprioceptive block from past frames. The
manifold-conditioned student augments this base observation with the
32-dimensional dynamic VAE latent
$\mathbf{z}_t(m)$, computed from the length-20 lookahead reference window
starting at the current reference frame:
\begin{equation}
  \mathbf{o}^{\mathrm{student}}_t \in
  \mathbb{R}^{69+32}=\mathbb{R}^{101}.
\end{equation}

The policy outputs a 12-dimensional scaled action $\mathbf{a}_t$, which is converted into joint-position offsets from the nominal standing pose:
\begin{equation}
  \mathbf{q}^{\mathrm{tar}}_t = \mathbf{q}_0 + \alpha\mathbf{a}_t,
  \qquad \alpha=0.25.
\end{equation}

\noindent\textbf{Model Architecture and Inference.}
Each per-motion specialist is a PPO actor with empirical observation
normalization followed by a three-layer MLP, which maps the
69-dimensional tracking observation to the action mean used for deterministic
expert queries.

The reference-window VAE used for manifold conditioning operates on normalized
20-frame kinematic reference windows. Its encoder is a two-layer MLP with
separate heads for $\boldsymbol{\mu}$ and $\log\boldsymbol{\sigma}^2$, producing
a 32-dimensional latent code, and its decoder is a mirrored two-layer MLP
that reconstructs the input window. 

The full-train flow policy student uses the transformer-history architecture.
The current 45-dimensional proprioceptive block, the 24-dimensional reference
command, the 32-dimensional VAE latent condition, and 10 past proprioceptive
frames are embedded as tokens and processed by a 4-layer Transformer encoder.
The flow velocity field conditions on the resulting observation embedding, the
current action sample, and the diffusion time, and predicts the 12-dimensional
velocity with a three-layer MLP. During rollout and evaluation, actions are
generated by integrating the learned velocity field from Gaussian action noise;
unless otherwise specified, all experiments use five ODE steps ($D=5$).

\noindent\textbf{Reward.}
We use a dense tracking reward for specialist and residual policy training:
\begin{equation}
  r_t = \sum_i w_i r^{(i)}_t.
\end{equation}
Each tracking term uses an exponential kernel
$r^{(i)}_t=\exp(-\|\mathbf{e}^{(i)}_t\|^2/\sigma_i^2)$ over pose or velocity
errors. Table~\ref{tab:motion_tracking_reward} summarizes the terms. The
residual policy additionally uses a calf-joint power-safety penalty to discourage
large negative mechanical power.

\begin{table}[!ht]
\centering
\small
\caption{Reward terms used for specialist and residual policy training. The
tracking terms use exponential kernels over pose and velocity errors, while the
penalty terms discourage non-smooth actions, joint-limit violations, and
self-collisions; the residual policy additionally uses a power-safety penalty
outside this table.}
\label{tab:motion_tracking_reward}
\begin{tabular}{lcc}
\toprule
\textbf{Reward Term} & \textbf{Weight} & \textbf{Formulation} \\
\midrule
\textit{Tracking terms} & & \\
\quad Global torso position & 0.5 & $\exp(-\|\mathbf{p}_{\mathrm{err}}\|^2/0.3^2)$ \\
\quad Global torso orientation & 0.5 & $\exp(-\|\boldsymbol{\theta}_{\mathrm{err}}\|^2/0.4^2)$ \\
\quad Relative body position & 3.0 & $\exp(-\|\mathbf{p}^{\mathrm{rel}}_{\mathrm{err}}\|^2/0.3^2)$ \\
\quad Relative body orientation & 1.0 & $\exp(-\|\boldsymbol{\theta}^{\mathrm{rel}}_{\mathrm{err}}\|^2/0.4^2)$ \\
\quad Body linear velocity & 1.0 & $\exp(-\|\mathbf{v}_{\mathrm{err}}\|^2/1.0^2)$ \\
\quad Body angular velocity & 1.0 & $\exp(-\|\boldsymbol{\omega}_{\mathrm{err}}\|^2/3.14^2)$ \\
\midrule
\textit{Penalty terms} & & \\
\quad Action rate & $-0.1$ & $\|\mathbf{a}_t-\mathbf{a}_{t-1}\|^2_2$ \\
\quad Joint limit & $-10.0$ & joint-limit violation count \\
\quad Self-collision & $-0.1$ & undesired contact count \\
\bottomrule
\end{tabular}
\end{table}

\noindent\textbf{Domain Randomization and Noise.}
To reduce the sim-to-real gap, we expose policies during training to sensing noise, state-estimation errors, external disturbances, and moderate physical-parameter variation. Specifically, we inject actor observation noise into joint positions, joint velocities, base angular velocity, and orientation, while keeping critic observations clean. Periodic base pushes and randomization over friction, center-of-mass offsets, and PD gain scales further improve robustness. We also use reference-state initialization (RSI) \cite{liao2025BeyondMimicMotionTracking}, resetting the robot around states sampled from the target reference, including root position, root orientation, root velocity, joint positions, and motion start phase. RSI prevents training from repeatedly starting near the same standing or early-reference states and forces the tracker to recover from off-reference states throughout the motion.

\subsection{Locomotion}
\label{sec:locomotion}


Velocity-driven locomotion, autonomously synthesizing stable gaits from velocity commands $(v_x, v_y, \dot{\psi})$ without explicit kinematic templates, is the most fundamental capability for quadruped deployment, underpinning autonomous navigation, human following, and teleoperation. While standard flat-ground trotting has been well addressed by RL with heuristic reward shaping, practical deployment demands two capabilities beyond basic mobility that remain largely unsolved.

\textbf{Biomimetic gait with velocity-tracking precision.}\quad In human-centric scenarios such as home companionship and indoor service, rigid ``robotic'' gaits are both energy-inefficient and socially unacceptable---smooth, animal-like movements are essential for interaction safety and user comfort. Adversarial imitation learning (e.g., AMP~\cite{peng2021amp}) can reproduce such gaits from animal motion data, but suffers from mode collapse on heterogeneous datasets, unbounded rewards that destabilize training, and severe heading drift under out-of-distribution (OOD) directional commands. The core challenge is: \emph{how to retain biomimetic naturalness without sacrificing precise velocity tracking?}

\textbf{All-terrain adaptability.}\quad Outdoor deployment---industrial inspection, last-mile delivery, search and rescue---requires traversing stairs, gaps, and overhanging obstacles. This demands 3D visual perception in the control loop, yet end-to-end RL from point clouds is prohibitively sample-inefficient, and 2.5D height maps fail to capture the full 3D topology for proactive obstacle avoidance.

We first establish a shared \emph{robust baseline} (Section~\ref{sec:robust_loco}) that provides sim-to-real fidelity and hardware safety through decoupled representation learning and constrained optimization. Built upon this foundation, we develop two specialized locomotion policies tailored to these complementary scenarios:
\begin{itemize}
    \item \textbf{Biomimetic Gait with Omnidirectional Ability} (Section~\ref{sec:biomimetic}): a bounded conditional diffusion prior with omnidirectional command-correction produces naturalistic, energy-efficient gaits---walk, trot, and bound transition fluidly with speed---while maintaining high-precision velocity tracking across all directions;
    \item \textbf{All-Terrain Locomotion} (Section~\ref{sec:perceptive}): a three-stage privileged-to-perceptive framework with temporal LiDAR memory and terrain-predictive supervision enables robust all-terrain traversal on complex unstructured environments.
\end{itemize}


\subsubsection{Robust Baseline}
\label{sec:robust_loco}

Real-world locomotion is a Partially Observable MDP: payload variations, CoM shifts, and terrain friction are unobservable from proprioception alone. Deploying unconstrained actions further risks exceeding motor limits. We address both challenges through three complementary mechanisms.

\paragraph{Self-Supervised Implicit Representation Learning}
We learn an implicit proprioceptive latent with a Barlow Twins-style temporal consistency objective~\cite{zbontar2021barlow}. From the 10-step proprioceptive buffer, the actor masks privileged base linear velocity and constructs two neighboring 5-step views: the latest history window and a one-step-shifted window formed by appending the current observation. A shared history encoder and projector map both views to features, whose cross-correlation matrix \(C\) is regularized toward the identity:
\[
\mathcal{L}_{\mathrm{BT}}
=
\sum_i (1-C_{ii})^2
+
\lambda_{\mathrm{off}}
\sum_{i\neq j} C_{ij}^2,
\quad
\lambda_{\mathrm{off}}=5\times10^{-3}.
\]
This aligns corresponding feature dimensions while suppressing redundancy, yielding a compact history-dependent proprioceptive representation without negative samples.

\paragraph{Explicit State Estimation}
We append \emph{explicit regression heads} to the representation layer, supervised by privileged simulation data, to estimate the robot's true base linear velocity, payload mass, and CoM coordinates. Coupled with a gravity-gated reward mechanism, the robot perceives its physical state online, enabling adaptive stiffness modulation under uncalibrated loads or external disturbances.

\paragraph{Hardware-Safe RL via Constrained MDP (NP3O)}
To eradicate safety hazards at the training source, we formulate the control domain as a Constrained Markov Decision Process (CMDP) optimized via \emph{Normalized Penalized PPO (NP3O)}. Unlike conventional penalty reward shaping, we define joint positions, velocities, and output torques as independent hard constraint boundaries:
\begin{equation}
    L_{\text{viol}} = \sum_{i=1}^{3} \lambda_i^{(t)} \max\!\left(0,\; C_{\text{surr}}^{(i)} + \tilde{v}_i\right),
\end{equation}
where $C_{\text{surr}}^{(i)}$ is the surrogate cost objective evaluated using normalized cost advantages, $\tilde{v}_i$ is the constraint margin term, and $\lambda_i^{(t)}$ is an exponentially increasing penalty coefficient. Owing to the ReLU gating, the penalty contributes no gradient when $C_{\text{surr}}^{(i)} + \tilde{v}_i \leq 0$, preserving ample exploration space in early training while strictly suppressing constraint violations upon convergence. This achieves \emph{zero safety violations} even during extreme command executions on real hardware.

\subsubsection{Biomimetic Gait with Omnidirectional Ability}
\label{sec:biomimetic}

Building upon the robust baseline, we elevate locomotion from mechanically functional to biomimetically natural. GAN-based adversarial imitation (e.g., AMP) suffers from three bottlenecks on heterogeneous datasets: (1) mode collapse, (2) unbounded adversarial rewards destabilizing training, and (3) forward bias causing heading drift under OOD commands. We address all three through \emph{Diff-CAST} (Diffusion-guided Constraint-Aware Symmetric Tracking)\cite{diffcast}.

\paragraph{Action-Agnostic Diffusion Prior (CC-Diffusion)}
We replace the GAN discriminator with a \emph{command-conditioned diffusion state-transition model}. Instead of evaluating state-action pairs $(s_t, a_t)$, requiring torque data impractical from biological motion capture, we model temporal state transitions $x_t = (s_t, s_{t+1})$. This action-agnostic formulation decouples stylistic learning from the actuator domain, enabling cross-embodiment retargeting.

The model is conditioned on a domain concept $c \in \{c^+, c^-\}$ (expert vs.\ agent) and the velocity command $v^{\text{cmd}} \in \mathbb{R}^3$ via early fusion. We compute the conditional denoising MSE under both hypotheses:
\begin{align}
    L^+(x_t) &= \mathbb{E}_{k, \varepsilon}\left[\|\varepsilon - \varepsilon_\varphi(x_{t,k}, c^+, k)\|^2\right], \\
    L^-(x_t) &= \mathbb{E}_{k, \varepsilon}\left[\|\varepsilon - \varepsilon_\varphi(x_{t,k}, c^-, k)\|^2\right],
\end{align}
and analytically derive a classification probability via softmax:
\begin{equation}
    D_\varphi(x_t) = \frac{\exp(-L^+(x_t))}{\exp(-L^+(x_t)) + \exp(-L^-(x_t))}.
\end{equation}

By conditioning on velocity commands, the diffusion prior evaluates each step against the aligned expert sub-manifold, eliminating the command-override dilemma of unconditioned priors.

\paragraph{Bounded Stylistic Reward}
Standard adversarial rewards use an unbounded log-likelihood ratio $r = \log(D) - \log(1-D)$, where early-exploration deviations trigger extreme spikes ($r \to -\infty$) that destabilize the PPO value network. We instead use the bounded classification probability directly:
\begin{equation}
    r_{\text{diff}} = D_\varphi(x_t) \in [0, 1].
\end{equation}
This retains mode-seeking gradient guidance while preventing extreme spikes, enabling zero-shot fluid transitions among walk, trot, and bound gaits.

\paragraph{Omnidirectional Velocity Tracking via SACC}
Biological datasets are intrinsically forward-biased, causing unconditioned priors to override lateral/backward commands with persistent heading drifts. We propose Symmetric Augmented Command Construction (SACC), injecting geometric regularization at both data and architecture levels:

\begin{itemize}
    \item \textbf{Kinematic Symmetry.} A sagittal-plane mirror operator $M(\cdot)$ swaps contralateral legs and negates asymmetric spatial components (lateral velocity, roll, yaw rate). Augmenting the offline dataset with $M(\cdot)$ balances unilateral biases while a structural mirror symmetry loss on the actor-critic networks prevents unilateral limping:
    \begin{equation}
        L_{\text{sym}} = \lambda \left(\left\|\pi_\theta(s_t, v^{\text{cmd}}) - M_a\!\left(\pi_\theta(\tilde{s}_t, \tilde{v}^{\text{cmd}})\right)\right\|^2 + \left\|V_\Psi(s_t, v^{\text{cmd}}) - V_\Psi(\tilde{s}_t, \tilde{v}^{\text{cmd}})\right\|^2\right),
    \end{equation}
    where $\tilde{s}_t = M_s(s_t)$ and $\tilde{v}^{\text{cmd}} = M_v(v^{\text{cmd}})$ are the mirrored state and command.

    \item \textbf{Yaw Invariance.} To prevent the discriminator from acting as a directionally biased prior, heading-dependent planar features are dynamically rotated by a uniformly sampled yaw offset $\delta \sim U(-\pi, \pi)$ during diffusion updates, synthetically generating $360^\circ$ rotational data augmentations.
\end{itemize}

Co-optimizing these mechanisms decouples the biomimetic prior from velocity tracking at the feature level, driving heading deviation close to $0^\circ$ across all directions.

\subsubsection{All-Terrain Locomotion}
\label{sec:perceptive}
\begin{figure}[t]
\centering
\includegraphics[width=\linewidth]{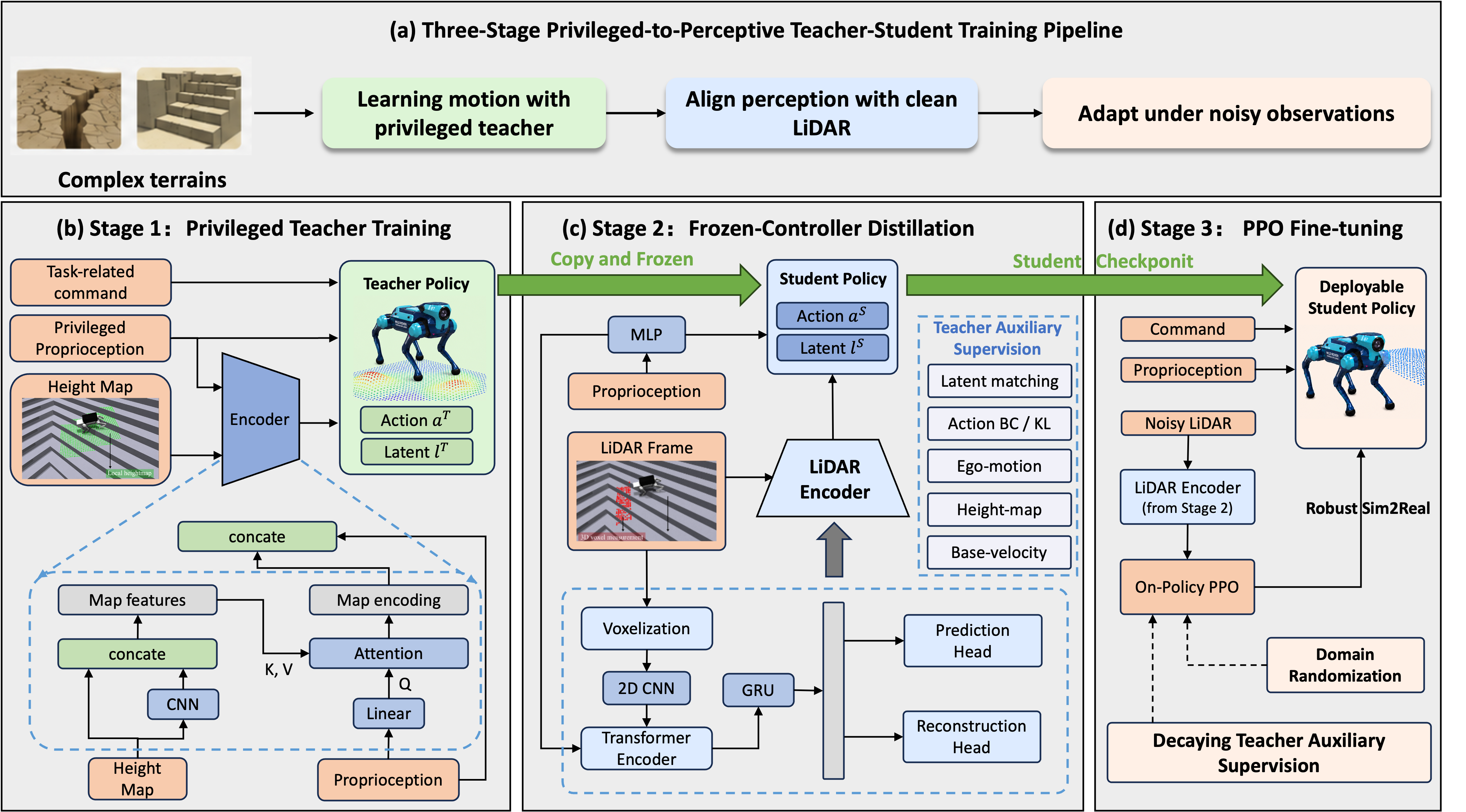}
\caption{Three-stage privileged-to-perceptive training pipeline for LiDAR-based all-terrain locomotion.
A privileged teacher is first trained with clean height and dynamics information.
The student then learns a LiDAR memory representation through frozen-controller distillation,
aligning its latent and action distribution with the teacher under clean LiDAR observations.
Finally, the distilled student is fine-tuned with on-policy PPO under noisy, domain-randomized
LiDAR inputs, while decaying teacher regularization preserves stable locomotion behavior.}
\label{fig:training_pipeline}
\end{figure}

To enable robust locomotion over complex terrains, we adopt a three-stage
privileged-to-perceptive teacher-student training framework, as shown in
Fig.~\ref{fig:training_pipeline}. The framework follows privileged teacher training $\rightarrow$ clean LiDAR memory
distillation $\rightarrow$ noisy on-policy student PPO fine-tuning.
Instead of directly training a LiDAR policy from scratch, the student first
learns to recover the teacher's privileged terrain and dynamics latent through
temporal LiDAR memory, and is then fine-tuned under deployable noisy
observations.
The training setup uses two groups of information:
\begin{itemize}[leftmargin=1.2em]
  \item \emph{Teacher privileged observations}: clean proprioception, height map, base velocity, contact states, and dynamics variables such as friction, mass, CoM offset, external pushes, and joint stiffness/damping.
  \item \emph{LiDAR domain randomization}: sensor noise, point dropout, holes, pose perturbation, extrinsic disturbance, and scan delay.
\end{itemize}

\paragraph{Privileged teacher training}
We first train a teacher policy using privileged observations and dynamics-related
variables from the setup above. The teacher is trained on diverse
terrains with curriculum learning and domain randomization, enabling robust
obstacle-traversing behaviors. It later serves as both a strong locomotion
controller and a privileged reference for student training.

\paragraph{Clean LiDAR memory distillation}
In the second stage, a student is distilled from the teacher using clean LiDAR
memory observations. The student actor and action distribution are initialized
from the teacher and kept frozen, while only the LiDAR memory encoder and
auxiliary heads are optimized. This separates perceptual representation learning
from locomotion control, avoiding degradation of the teacher's learned controller.
The student takes an 8-frame LiDAR memory as input. Each scan is voxelized
into a body-frame 3D occupancy grid with scan-age information. A temporal encoder
combines per-frame CNN features, scan age, and learned ego-motion compensation
through a GRU. The resulting latent is supervised by teacher-guided matching and
auxiliary predictions, including base velocity, ego-motion, and terrain geometry.
These objectives encourage the latent representation to capture control-relevant,
terrain-aware, and motion-aware information.

\paragraph{Noisy on-policy student PPO fine-tuning}
Finally, the distilled student is fine-tuned with PPO using deployable
observations, including commands, proprioception, and noisy LiDAR memory. During
rollout, the student executes its own actions, preserving the on-policy nature of
PPO and enabling closed-loop adaptation under perceptive observations. LiDAR
domain randomization from the setup above is gradually introduced through a curriculum.
To stabilize fine-tuning, teacher regularization terms such as behavior cloning,
action-distribution KL, and latent matching are retained at the beginning and
then gradually decayed. This allows the student to improve task performance under
its own closed-loop behavior while remaining close to the teacher's stable control
manifold.

Overall, the proposed training pipeline combines privileged teacher learning,
frozen-controller LiDAR memory distillation, and regularized on-policy PPO
fine-tuning. It improves perceptive traversal robustness and safety on complex
terrains, at the cost of slightly higher energy consumption due to more cautious
foothold selection.

\subsection{Scene Interaction}
\label{subsec:scene-interaction}

Scene interaction lies at the intersection of quadruped human--robot interaction, legged loco-manipulation, and language-conditioned robot control. Recent quadruped HRI studies investigate intuitive speech and gesture interfaces for robot dogs, indicating the need for interaction mechanisms beyond joystick-style locomotion commands~\cite{shin2024NonverbalInteractionInterface}. In social robotics, handshaking has also been studied as a representative contact-rich greeting behavior, where the robot must reach toward a human partner and respond to physical interaction~\cite{prasad2021LearningHumanlikeHand}. These works motivate contact-oriented interaction as a meaningful target capability, but they do not directly solve whole-body quadruped control under real-time perception and balance constraints.

Related legged loco-manipulation methods address physical interaction by explicitly coupling balance, contact forces, and whole-body motion~\cite{sombolestan2023HierarchicalAdaptiveLocomanipulation,liu2024VisualWholebodyControl}. In parallel, language-conditioned robotic systems and VLA models have shown how high-level instructions can be grounded into robot skills or end-to-end manipulation policies~\cite{ahn2022CanNotSay,driess2023PaLMEEmbodiedMultimodal,zitkovich2023RT2VisionlanguageactionModels,kim2024OpenVLAOpenSourceVisionLanguageAction,zhao2023ChatEnvironmentInteractive,liu2026comatrack}. However, applying a monolithic vision-to-action policy to contact-oriented quadruped interaction remains difficult because perception uncertainty, approach behavior, contact timing, balance, and compliance must be solved simultaneously. We therefore formulate scene interaction as a compositional control problem that reuses locomotion for global approach and motion tracking for local contact execution, while introducing lightweight perception and IK-generated references to connect scene targets to executable quadruped behaviors.

\subsubsection{Hand-Shaking as a Case Study}
\label{subsubsec:hand-shaking-study}

The scene-interaction capability is instantiated in this report through a hand-shaking case study: the robot approaches a detected human hand, reaches with a front leg, responds compliantly to human motion, and returns to locomotion. Compared with free-space motion, this task requires low-latency target updates, safe contact handling, and controllable behavior during both approach and physical interaction. These requirements are challenging because human motion is intermittent and partially occluded, perception can be delayed or noisy, and the controller must remain stable while reacting to contact.

We therefore adopt a separated perception-and-control setting: open-source vision models and depth sensing estimate the hand target, while the controller reuses existing locomotion and motion-tracking modules to preserve predictable command interfaces, controllability, and extensibility. Compared with a monolithic end-to-end RL policy, this decomposition reduces dependence on large-scale vision-based teleoperation data and avoids coupling approach, balance, leg lifting, and contact-pose objectives into a single reward. New task data are introduced only at the control level through IK-generated reaching references for the local hand-shaking behavior.\footnote{End-to-end visual RL is especially sensitive to delayed or biased hand detections at deployment; without substantial realistic perception-conditioned training data, these errors can create a large sim-to-real gap and make behaviors such as approach speed and final touch pose difficult to control.}

\subsubsection{Methodology}
\label{subsubsec:scene-pipeline}

\begin{figure}[t!]
    \centering
    \includegraphics[width=\textwidth]{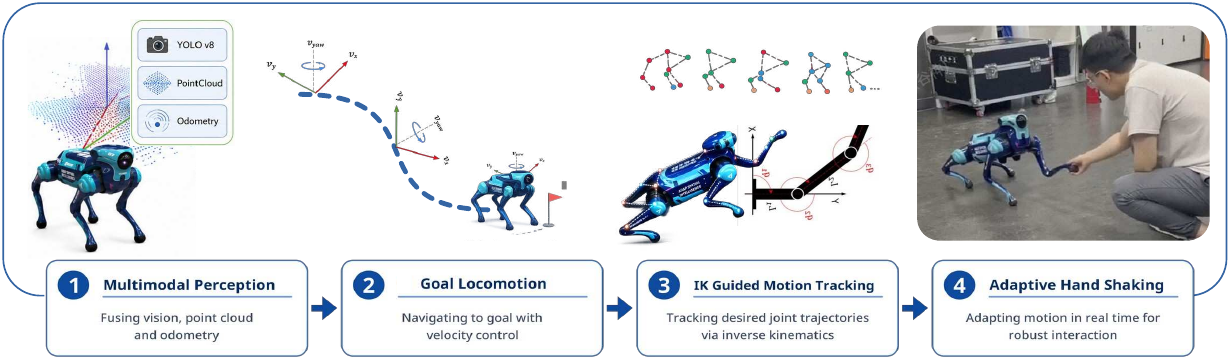}
    \caption{Hand-shaking case-study pipeline for scene interaction. The task integrates locomotion for approach and motion tracking for local execution, with IK-generated references providing the reaching and hand-shaking data.}
    \label{fig:scene}
\end{figure}

The interaction follows a perception--locomotion--IK--motion-tracking pipeline (see Figure~\ref{fig:scene}). The perception module estimates the hand position from camera hand landmarks and LiDAR depth, while odometry helps maintain a stable robot-relative target during approach.

\paragraph{Goal-directed Locomotion} Let $p_h^B=(x_h,y_h,z_h)$ denote the detected hand position in the robot base frame. During approach, the hand target is converted into a locomotion command $u_{\mathrm{loc}}=(v_x,v_y,\dot\psi)$ by
\begin{equation}
  u_{\mathrm{loc}} =
  \left[
  k_{xy}(x_h-d_x),\quad
  k_{xy}y_h,\quad
  k_{\psi}\,\mathrm{atan2}(y_h,\max(x_h,\epsilon))
  \right]^\top,
\end{equation}
where $d_x$ is the desired forward standoff and $\epsilon$ prevents a singular heading estimate. The resulting forward, lateral, and yaw-rate commands are clipped to their configured limits before being sent to the locomotion policy (c.f. Section~\ref{sec:locomotion}).

\paragraph{IK-guided Motion Tracking} After the robot stops near the person, IK uses the stabilized hand target to generate a front-leg reaching trajectory from the standing pose to the hand-shaking pose and back. These IK trajectories become reference data for motion-tracking training, because IK is kinematic and does not solve balance or support dynamics. During deployment, the system blends from locomotion to motion tracking, executes the hand-shaking reference, detects completion, and blends back to locomotion. The IK reference is generated by solving a weighted whole-body objective,
\begin{equation}
  q^{\mathrm{IK}}_{1:T}
  = \arg\min_{q_{1:T}} \sum_{t=1}^{T}
  \left(
  \|W_F(p_F(q_t)-p_h^B)\|_2^2
  + \sum_{i\in\mathcal{S}} w_i^2\|p_i(q_t)-p_i^0\|_2^2
  + \|W_q(q_t-q^0)\|_2^2
  \right),
\end{equation}
where $p_F$ is the active front-foot endpoint, $\mathcal{S}$ denotes support feet and base sites that should remain near their nominal positions, and $q^0$ is the nominal posture. 
The motion-tracking policy (c.f. Section~\ref{sec:motion_track}) is then trained to execute the IK reference dynamically. 

\paragraph{Adaptive Interaction} During shaking, the active thigh and calf gains are reduced to a minimal support level so the robot can respond compliantly to human motion while maintaining posture. Completion is based on a proprioceptive external-stimulus score,
\begin{equation}
  s(t)=\max_{j\in\mathcal{A}} |\dot q_j(t)-\dot q_j^{\mathrm{cmd}}(t)|,
\end{equation}
where $\mathcal{A}$ contains the active thigh and calf joints. After a warm-up interval, sustained above-threshold stimulus marks the interaction as engaged; a later quiet interval triggers the return segment. Timeout and safety guards remain active throughout the task.  

\section{System}
\label{sec:system}
\FloatBarrier

This section describes the deployment system that connects the learned ABot-C0 motion policies to the physical quadruped platform. We first summarize the hardware and compute configuration that supports perception, audio interaction, onboard control, and cloud-side reasoning. We then describe the layered runtime stack that coordinates locomotion, motion tracking, and task-level interaction through a shared robot I/O interface.

\subsection{Hardware Configuration}
\label{subsec:system-hardware}
\begin{figure}[htbp]
    \centering
    \includegraphics[width=0.92\textwidth]{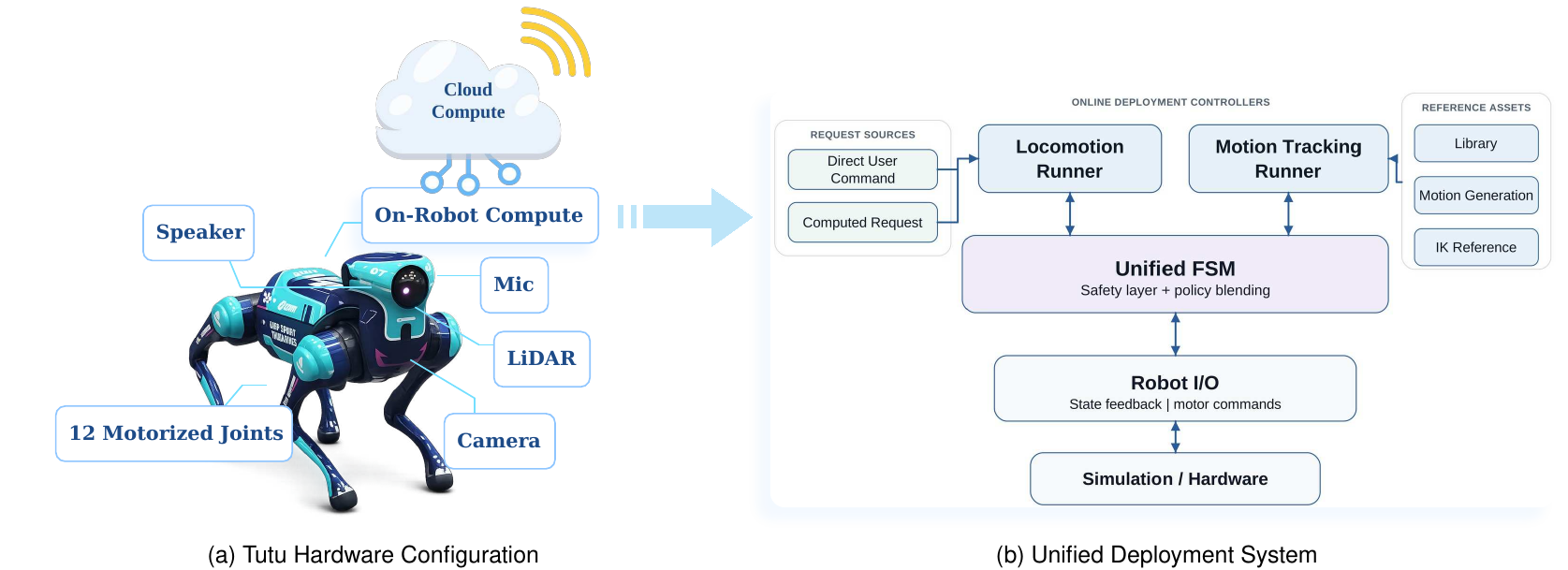}
  \caption{Hardware configuration and deployment system of ABot-C0. (a) The Tutu hardware platform integrates multi-view cameras, LiDAR, audio modules, onboard edge computing, and cloud-side reasoning support for real-world perception and interaction. (b) The deployment stack is hosted by the on-robot compute module highlighted in (a) and follows a layered design: task-level commands are coordinated across resident locomotion and motion-tracking runners, then passed through shared command synthesis, safety checks, and robot I/O.}
  \label{fig:deploy}
\end{figure}

As shown in Fig.~\ref{fig:deploy}(a), our quadruped platform \textbf{Tutu} is equipped for multimodal perception, onboard control, and high-level interaction. Three SENSING cameras cover the front, left, and right views with a $120^{\circ}\times90^{\circ}$ field of view, while a 96-channel RoboSense Airy LiDAR provides dense spatial sensing for navigation and interaction. An AISpeech microphone-speaker module supports voice interaction. Real-time onboard computation runs on an NVIDIA Jetson AGX Orin 64GB module, and computationally heavier reasoning modules can be offloaded to cloud infrastructure when needed.

\FloatBarrier

\subsection{Runtime Architecture}
\label{subsec:system-runtime}

As shown in Fig.~\ref{fig:deploy}(b), the runtime stack is hosted by the on-robot compute module in Fig.~\ref{fig:deploy}(a) and keeps the locomotion and motion-tracking runners alive in the background. The locomotion runner continuously accepts velocity commands for standing, walking, approach, and retreat. The motion-tracking runner maintains the context needed to execute reference motions from a motion library, a motion generation model, or an IK-generated interaction trajectory. Keeping both runners resident avoids reloading policies or rebuilding observations when a task switches mode, which reduces latency during interaction.
The coordination layer decides which runner output is active and blends motor targets when switching between policy outputs or stored joint postures. The state machine is therefore used mainly for arbitration, smooth transition, and safety gating.

Motion controller operates at the joint-command level. Learned policy actions are converted into desired joint positions, stiffness, damping, and optional feedforward torques, while feedback contains joint state, inertial measurements, estimated torque, and actuator temperature. For position-control policies, scaled actions define target joint positions $\mathbf{q}^{\mathrm{tar}}_t$, and the low-level command path applies per-joint PD control, $\boldsymbol{\tau}_t=\mathbf{k}_p\odot(\mathbf{q}^{\mathrm{tar}}_t-\mathbf{q}_t)-\mathbf{k}_d\odot\dot{\mathbf{q}}_t$, before command publication. Hardware and MuJoCo simulation use the same robot-state and motor-command abstraction, so the deployment controller can be tested in simulation before hardware execution. The motor-command loop runs at 200\,Hz, while the decision tick, locomotion inference, and motion-tracking inference run at 50\,Hz; the latest runner outputs are consumed by the high-rate command loop without blocking command publication.

\section{Experiments}
\label{sec:experiments}

This section evaluates the core motion capabilities of ABot-C0 across large-scale simulation (IsaacGym, IsaacSim, Mujoco) and physical deployment on our \textbf{Tutu} robot. 
We first analyze generalist motion tracking (Section~\ref{sec:motion_tracking_eval}), including specialist distillation, data scaling, motion curation, and manifold-conditioned references. We then evaluate locomotion robustness, biomimetic gait control, and perceptive all-terrain traversal (Section~\ref{sec:locomotion_exp}). Finally, we assess the hand-shaking scene-interaction case study (Section~\ref{subsec:scene-interaction-experiments}), focusing on IK reference quality and motion-tracking execution accuracy.

\subsection{Motion Tracking}
\label{sec:motion_tracking_eval}

\subsubsection{Experimental Setup}
\label{sec:motion_eval_setup}
The full training pool contains 7,076 reference motions for our quadruped
platform, collected from retargeted teleoperation recordings, designer-authored motions, and video-generated motions. We evaluate policies
on the training motions to measure seen tracking fidelity and on a held-out set
of 1,000 references to measure unseen generalization. Unless otherwise stated,
each reference motion is evaluated with five rollouts and flow-policy inference
uses five ODE steps. The default full-scale flow-policy setting uses a
transformer-history student with a history length of 10 and 30,000 training
iterations; deviations from this setting are specified in the corresponding
experiment.

We use mean per-joint position error (MPJPE) as the primary tracking metric.
For a rollout of length $T$ with $J$ tracked joints, MPJPE is computed as
\begin{equation}
  \mathrm{MPJPE}
  =
  \frac{1}{TJ}
  \sum_{t=1}^{T}\sum_{j=1}^{J}
  \left\|
    \mathbf{q}^{\mathrm{robot}}_{t,j}
    -
    \mathbf{q}^{\mathrm{ref}}_{t,j}
  \right\|_2,
\end{equation}
and reported in millimeters. Here $T$ is the rollout horizon, $J$ is the number
of tracked joints, and $\mathbf{q}^{\mathrm{robot}}_{t,j}$ and
$\mathbf{q}^{\mathrm{ref}}_{t,j}$ denote the Cartesian positions of the
$j$-th robot and reference joint at timestep $t$, respectively. We additionally
report success rate as the fraction of evaluation rollouts that complete the
assigned reference:
\begin{equation}
  \mathrm{Success\ Rate}
  =
  \frac{1}{N}
  \sum_{i=1}^{N}\mathbb{I}\left[\tau_i\ \mathrm{completes}\right],
\end{equation}
where $N$ is the number of evaluated rollouts, $\tau_i$ denotes the $i$-th
rollout trajectory, and $\mathbb{I}[\cdot]$ is the indicator function. A rollout
is counted as complete if it reaches the end of the assigned motion without
triggering an early termination.

\subsubsection{Specialist-to-Generalist Tracking Baseline}
\label{sec:motion_tracking_baseline}

\begin{table}[htbp]
\centering
\small
\caption{Specialist-to-generalist tracking baseline on 1,000 training motions. Values are MPJPE in mm and success rate; lower MPJPE and higher success are better, and bold numbers mark the best generalist result in each column.}
\label{tab:tracking_baseline_1k}
  \begin{tabular}{lcccc}
  \toprule
  \textbf{Method} & \multicolumn{2}{c}{\textbf{Seen}} & \multicolumn{2}{c}{\textbf{Unseen}} \\
  \cmidrule(lr){2-3}\cmidrule(lr){4-5}
  & \textbf{MPJPE ($\downarrow$ mm)} & \textbf{Success rate ($\uparrow$)} & \textbf{MPJPE ($\downarrow$ mm)} & \textbf{Success rate ($\uparrow$)} \\
  \midrule
  Specialist & 11.66 & 97.34\% & -- & -- \\
  \midrule
  Multi-motion RL & 22.18 & 87.24\% & 22.20 & 84.86\% \\
  Flow policy & 12.04 & 95.26\% & 16.51 & 86.42\% \\
  Flow policy + Residual & \textbf{11.98} & \textbf{95.49\%} & \textbf{16.50} & \textbf{86.43\%} \\
  \bottomrule
\end{tabular}
\end{table}

We first evaluate whether the specialist-to-generalist pipeline can produce a
strong generalist tracker from per-motion supervision. Table~\ref{tab:tracking_baseline_1k}
compares four policies trained on 1,000 motions. The \textit{Specialist}
baseline trains an independent PPO expert for each training motion, i.e., one dedicated policy model per reference, and therefore
serves as a seen-motion upper reference rather than a deployable generalist.
In contrast, \textit{Multi-motion RL} trains a single MLP policy from scratch with standard RL on the full multi-motion set, using the same model to track all references. \textit{Flow policy} distills the specialist ensemble into one
flow-matching policy using on-policy DAgger aggregation. \textit{Flow policy +
Residual} further freezes the distilled flow policy and learns a lightweight
residual correction with RL.

The results show that direct multi-motion RL is insufficient at this scale,
yielding 22.18 mm seen MPJPE and 22.20 mm unseen MPJPE, with substantially
lower seen success than the per-motion specialists. The flow policy closes most of
this gap while producing a single deployable policy: on seen motions it reduces
MPJPE from 22.18 to 12.04 mm and raises success from 87.24\% to 95.26\%.
Although the distilled policy remains below the per-motion specialist upper
reference (11.66 mm and 97.34\% seen success), it preserves most of the
specialist tracking fidelity while enabling generalist control. More
importantly, the flow policy substantially improves held-out performance, reducing
unseen MPJPE from 22.20 to 16.51 mm and increasing unseen success from 84.86\%
to 86.42\% relative to multi-motion RL. Residual RL provides a small but
consistent refinement, improving seen MPJPE from 12.04 to 11.98 mm and seen
success from 95.26\% to 95.49\%, while leaving unseen performance nearly
unchanged (16.51 to 16.50 mm MPJPE and 86.42\% to 86.43\% success). These
results indicate that specialist distillation into a flow policy supplies the
dominant generalist-tracking gain, whereas the residual layer primarily acts as
a local robustness correction.

\subsubsection{Flow Policy Scaling Law}
\label{sec:flow_policy_scaling}

We next examine whether quadruped motion tracking exhibits a data-scaling
effect under the same specialist-to-generalist flow policy pipeline. We vary
the number of training references from 30 motions to the full 7,076-motion pool
and evaluate each resulting policy on both seen and held-out unseen motions. At
each data scale, the seen evaluation set is identical to the corresponding
training-reference set, so its size follows the training-reference scale.
Table~\ref{tab:flow_scaling_law} reports one representative checkpoint at each
data scale.

\begin{table}[htbp]
\centering
\small
\setlength{\tabcolsep}{3pt}
\caption{Flow policy scaling law with respect to the training
motion volume. For each data scale, Seen is evaluated on the same motion dataset used for
training at that scale, matching the number of training samples. The seen-unseen MPJPE
gap is computed as Unseen MPJPE minus Seen MPJPE, and the success-rate gap is
computed as Unseen Success minus Seen Success in percentage points.}
\label{tab:flow_scaling_law}
\begin{tabular}{lcccccc}
\toprule
\textbf{Samples} & \multicolumn{2}{c}{\textbf{Seen}} & \multicolumn{2}{c}{\textbf{Unseen}} & \multicolumn{2}{c}{\textbf{Seen-Unseen Gap}} \\
\cmidrule(lr){2-3}\cmidrule(lr){4-5}\cmidrule(lr){6-7}
& \textbf{MPJPE ($\downarrow$ mm)} & \textbf{Success rate ($\uparrow$)} & \textbf{MPJPE ($\downarrow$ mm)} & \textbf{Success rate ($\uparrow$)} & \textbf{MPJPE (mm)} & \textbf{Success rate} \\
\midrule
30 & 14.44 & 92.00\% & 24.61 & 84.30\% & 10.17 & -7.70 pp \\
100 & \textbf{11.70} & \textbf{97.20\%} & 20.27 & 85.26\% & 8.58 & -11.94 pp \\
300 & 12.15 & 96.60\% & 18.84 & 85.90\% & 6.70 & -10.70 pp \\
1,000 & 12.04 & 95.26\% & 16.51 & 86.42\% & 4.47 & -8.84 pp \\
3,000 & 11.78 & 94.86\% & 15.15 & 88.22\% & 3.37 & -6.64 pp \\
full (7,076) & 12.38 & 92.74\% & \textbf{14.79} & \textbf{88.54\%} & \textbf{2.41} & \textbf{-4.20 pp} \\
\bottomrule
\end{tabular}
\end{table}


\begin{wrapfigure}{r}{0.52\linewidth}
    \vspace{-15pt} 
    \centering
    \includegraphics[width=0.98\linewidth]{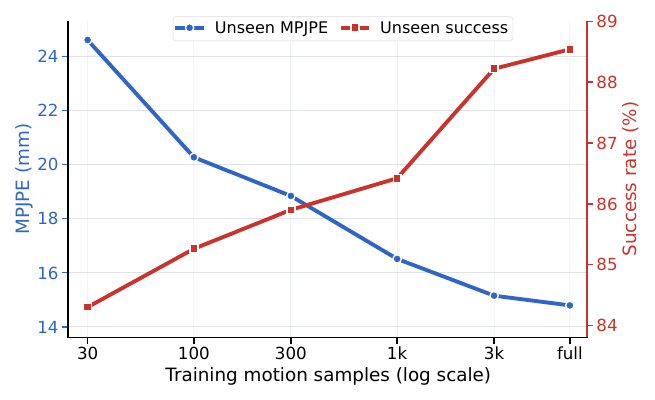}
    \caption{Unseen scaling trend for the flow policy. The figure visualizes the unseen metrics from Table~\ref{tab:flow_scaling_law} against the number of training motions. Larger motion libraries consistently reduce unseen MPJPE and improve unseen success rate.}
    \label{fig:motion_scaling_law_unseen}
    \vspace{-10pt} 
\end{wrapfigure}

The results reveal a clear scaling trend in generalization. Increasing the
training set from 30 to 7,076 motions reduces unseen MPJPE from 24.61 to
14.79 mm and improves unseen success from 84.30\% to 88.54\%. The full-data
policy also yields the smallest seen-unseen MPJPE gap, reducing the gap from
10.17 mm at 30 motions to 2.41 mm at 7,076 motions. Although the best seen-only
accuracy is achieved at smaller data scales, larger motion libraries
substantially improve held-out tracking and narrow the generalization gap. This
establishes a data-scaling law for quadruped motion tracking and motivates a
complementary question: \emph{under a fixed motion budget, can reference selection
produce better generalist policies than random subsampling?}

\subsubsection{Dynamic-Aware Motion Curation}
\label{sec:dynamic_aware_motion_curation_eval}

We next study whether a fixed-size training set can be made more effective by
curating references according to dynamic quality rather than sampling motions
uniformly at random. To efficiently isolate the effect of motion selection, all runs in this
subsection use 70\% of the full training pool and share the same lightweight
MLP flow policy without observation history. Table~\ref{tab:dynamic_aware_curation}
compares random subsampling, single-score curation ablations, and the combined
PRF-score selector.


Under the same 70\% motion budget, all score-based curation strategies reduce
unseen MPJPE relative to random subsampling, confirming that reference quality
predicts downstream tracking performance. Rollout-only curation is the strongest
single-score variant, while the combined PRF selector achieves the best overall
result: 11.17 mm / 94.81\% on seen motions and 15.56 mm / 85.66\% on unseen
motions. This suggests that physical feasibility, closed-loop executability,
and flow-policy confidence provide complementary signals for fixed-budget
reference selection.

\begin{table}[!ht]
\centering
\small
\setlength{\tabcolsep}{3pt}
\caption{Dynamic-Aware Motion Curation under a fixed 70\% motion budget. All
runs use the same lightweight history-free MLP flow policy and 10k training
iterations, so performance differences primarily reflect the selected motion
subset. Each cell reports MPJPE (mm) / success. The single-score rows isolate
Physical feasibility $p(m)$, Rollout executability $r(m)$, and Flow confidence $f(m)$, while
PRF-ranked curation uses $S_{\mathrm{cur}}(m)$ after complexity-bin
coverage. Score definitions are given in Section~\ref{sec:motion_methodology}.}
\label{tab:dynamic_aware_curation}
\begin{tabular}{lcc}
\toprule
\textbf{Sample Split} & \multicolumn{1}{c}{\textbf{Seen}} & \multicolumn{1}{c}{\textbf{Unseen}} \\
\cmidrule(lr){2-2}\cmidrule(lr){3-3}
& \textbf{MPJPE ($\downarrow$ mm) / Success rate ($\uparrow$)} & \textbf{MPJPE ($\downarrow$ mm) / Success rate ($\uparrow$)} \\
\midrule
random baseline & 14.01 / 88.47\% & 16.04 / 85.53\% \\
physical-only ($p(m)$) & 13.02 / 91.18\% & 15.77 / 85.55\% \\
rollout-only ($r(m)$) & 11.33 / 94.73\% & 15.66 / 85.02\% \\
flow-conf.-only ($f(m)$) & 11.99 / 94.03\% & 15.87 / 85.57\% \\
PRF-ranked ($S_{\mathrm{cur}}(m)$) & \textbf{11.17} / \textbf{94.81\%} & \textbf{15.56} / \textbf{85.66\%} \\
\bottomrule
\end{tabular}
\end{table}

\subsubsection{Manifold-Calibrated Reference Conditioning}
\label{sec:manifold_conditioning_eval}

We next evaluate whether exposing the student policy to learned
reference-manifold information improves generalist tracking. The comparison in
Table~\ref{tab:manifold_conditioning} keeps the training set, architecture, and
evaluation protocol fixed, and varies only the student observation.

\begin{table}[htbp]
\centering
\small
\setlength{\tabcolsep}{5pt}
\caption{MCRC under the default
full-scale flow-policy setting. $\mathbf{z}$ denotes the VAE
reference-window latent code, and $\mathbf{e}_{\mathrm{recon}}$ denotes the
corresponding VAE reconstruction error. Each
performance cell reports MPJPE (mm) / success.}
\label{tab:manifold_conditioning}
\begin{tabular}{lcc}
\toprule
\textbf{Student Observation} & \multicolumn{1}{c}{\textbf{Seen}} & \multicolumn{1}{c}{\textbf{Unseen}} \\
\cmidrule(lr){2-2}\cmidrule(lr){3-3}
& \textbf{MPJPE ($\downarrow$ mm) / Success rate ($\uparrow$)} & \textbf{MPJPE ($\downarrow$ mm) / Success rate ($\uparrow$)} \\
\midrule
$\mathbf{o}_{69}$ & 12.38 / 92.74\%  & 14.79 / 88.54\% \\
$\mathbf{o}_{69}\oplus\mathbf{e}_{\mathrm{recon}}$ & 12.24 / 93.48\% & 13.98 / 89.42\% \\
$\mathbf{o}_{69}\oplus\mathbf{z}$ & \textbf{11.77} / 94.16\% & \textbf{12.53} / \textbf{91.02\%} \\
$\mathbf{o}_{69}\oplus\mathbf{z}\oplus\mathbf{e}_{\mathrm{recon}}$ & 11.91 / \textbf{94.18\%} & 12.86 / 90.76\% \\
\bottomrule
\end{tabular}
\end{table}

Here $\mathbf{z}$ is the time-indexed latent code produced by the
reference-window VAE and represents the current lookahead segment on the
learned motion manifold, while $\mathbf{e}_{\mathrm{recon}}$ is the corresponding
reconstruction error and serves as a scalar reliability cue. The latent
code provides the strongest overall conditioning signal. Relative to the
base observation $\mathbf{o}_{69}$, adding $\mathbf{z}$ reduces seen MPJPE
from 12.38 to 11.77 mm and unseen MPJPE from 14.79 to 12.53 mm, while
increasing unseen success from 88.54\% to 91.02\%.

The reconstruction cue also improves over the base observation when used alone,
but its gains are smaller than those from the latent code. Combining
$\mathbf{z}$ and $\mathbf{e}_{\mathrm{recon}}$ gives the best seen success rate
(94.18\%), but is slightly weaker than $\mathbf{z}$ alone on seen MPJPE and on
both unseen metrics. Overall, $\mathbf{o}_{69}\oplus\mathbf{z}$ provides the
best trade-off across seen and unseen motions, supporting the value of MCRC as
a manifold-conditioning signal.

\subsection{Locomotion}
\label{sec:locomotion_exp}

We validate the progressive locomotion framework organized along three dimensions: robust baseline (Section~\ref{sec:exp_safety}), biomimetic gait with omnidirectional ability (Section~\ref{sec:exp_biomimetic}), and all-terrain locomotion (Section~\ref{sec:exp_perceptive}).

\subsubsection{Robust Baseline}
\label{sec:exp_safety}

We evaluate our robust locomotion baseline by examining two aspects: (1) the impact of our implicit-explicit state representation on control robustness under unmodeled physical shifts, and (2) the execution safety guaranteed by NP3O on real hardware.

\paragraph{Ablation on Implicit-Explicit State Representation}
To evaluate the contribution of the proposed state representation to downstream
control robustness, we compare four architectural variants under the same reward
design and NP3O training setup. During deployment, ground-truth base velocity and
privileged physical parameters are not provided to the actor.

All models are evaluated over diverse velocity commands and a grid of unmodeled
physical conditions, including payload additions ($0$, $2$, and $4~\mathrm{kg}$),
lateral CoM shifts from $-5~\mathrm{cm}$ to $+5~\mathrm{cm}$, and standardized
lateral impulses.

\begin{table}[htbp]
\centering
\small
\caption{Robustness evaluation under payload variations, CoM shifts, and lateral perturbations.}
\label{tab:state_ablation}
\begin{tabular}{lccc}
\toprule
\textbf{Method} & \textbf{Vel. Err. (m/s)} $\downarrow$ & \textbf{Fall Rate (\%)} $\downarrow$ & \textbf{Rec. Time (s)} $\downarrow$ \\
\midrule
Current-only (No History) & 0.22 & 7.5 & 1.12 \\
MLP History Encoder       & 0.19 & 1.5 & 0.98 \\
Barlow Latent Only        & 0.16 & 0   & 0.86 \\
\textbf{Ours (Barlow + Explicit Estimates)} & \textbf{0.14} & \textbf{0} & \textbf{0.82} \\
\bottomrule
\end{tabular}
\end{table}

As shown in Table~\ref{tab:state_ablation}, temporal representations
substantially improve robustness under unmodeled payload, CoM, and impulse
perturbations. The current-only policy exhibits the highest tracking error and a
$7.5\%$ fall rate, while history-based models reduce failures and improve
recovery. In particular, Barlow-regularized latent features achieve zero observed
falls, indicating improved robustness to dynamics variations. Our full model
further combines these implicit temporal features with explicit estimates of
base velocity, mass, and CoM shift, achieving the lowest velocity error
($0.14~\mathrm{m/s}$), fastest recovery time ($0.82~\mathrm{s}$), and zero
observed falls. These results suggest that implicit and explicit state
representations are complementary for robust locomotion control.

\paragraph{Hardware Safety Constraints via NP3O}
We next evaluate the hardware safety of our Constrained Markov Decision Process (CMDP) framework. Policies trained purely with penalty-based reward shaping may still produce actuator commands that exceed hardware limits during aggressive maneuvers. To assess this behavior, we conduct a high-speed sprint stress test on the Tutu with a target velocity of $3.0~\mathrm{m/s}$. During the experiment, we record joint torque and velocity statistics and compare our NP3O controller with an unconstrained vanilla PPO and a penalty-based PPO.

\begin{table}[htbp]
\centering
\small
\caption{Hardware safety evaluation during high-speed locomotion ($3.0~\mathrm{m/s}$ sprint).}
\label{tab:safety}
\begin{tabular}{lcccc}
\toprule
\textbf{Method} & \textbf{Torque Viol.} $\downarrow$ & \textbf{Joint Vel. Viol.} $\downarrow$ & \textbf{Fall Rate (\%)} $\downarrow$ \\
\midrule
Vanilla PPO (Unconstrained) & 124.8 & 318.5 & 55.0 \\
Penalty-based PPO & 17.2 & 54.6 & 15.0 \\
\textbf{Ours (NP3O)} & \textbf{0} & \textbf{0} & \textbf{0} \\
\bottomrule
\end{tabular}
\end{table}

As shown in Table~\ref{tab:safety}, the unconstrained PPO frequently exceeds actuator limits, while the penalty-based PPO reduces but does not eliminate constraint violations. In contrast, NP3O strictly enforces the predefined hardware limits, achieving zero torque and joint velocity violations throughout the experiment.

\subsubsection{Biomimetic Gait with Omnidirectional Ability}
\label{sec:exp_biomimetic}

We evaluate Diff-CAST from two aspects: gait diversity and omnidirectional
command tracking. We briefly summarize the key results below, more detailed
experimental evaluations can be found in our paper~\cite{diffcast}.

\paragraph{Motion Diversity and Gait Realism}
We evaluate motion quality using FGD and visualize the learned semantic skill
manifold with UMAP.

\begin{figure}[!htbp]
    \centering
    \includegraphics[width=0.74\linewidth]{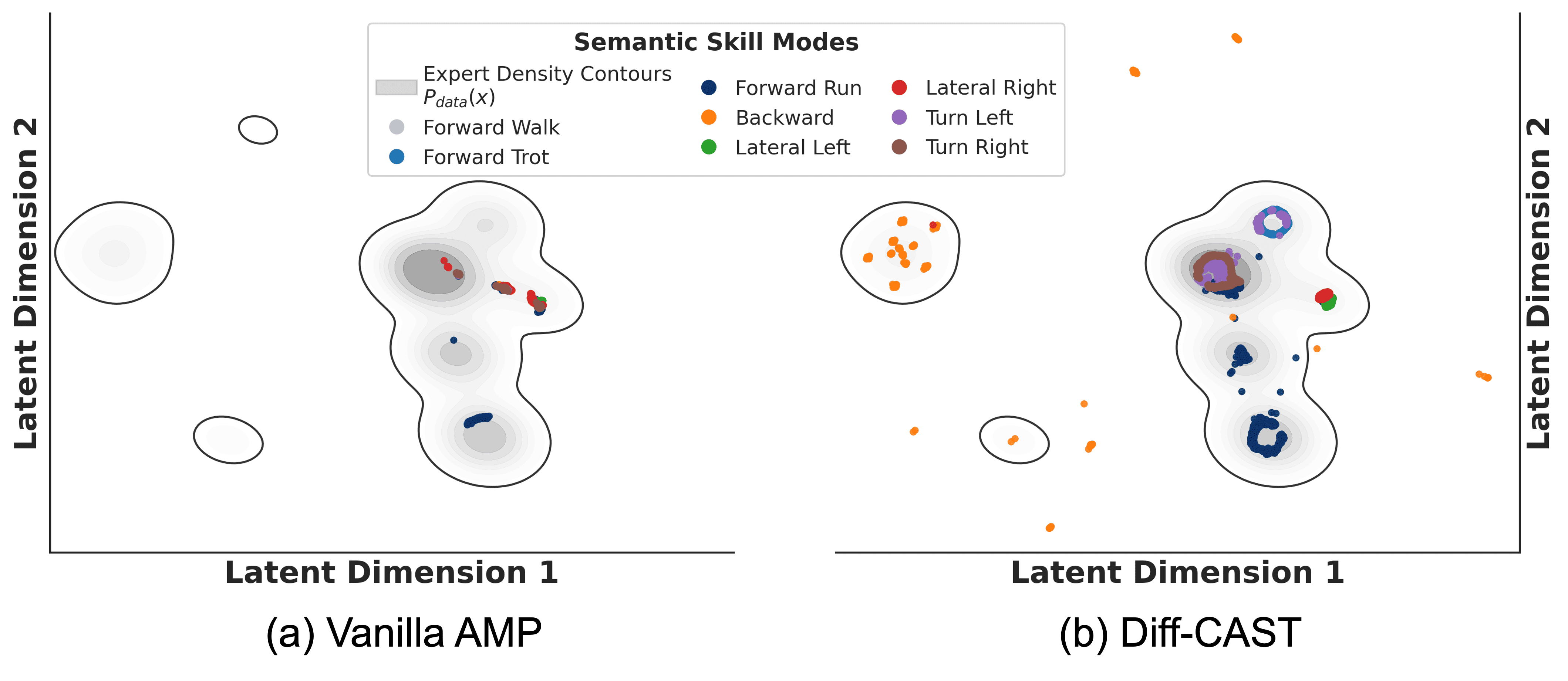}
    \caption{Latent dimension analysis via UMAP. Vanilla AMP (a) exhibits severe mode collapse, whereas Diff-CAST (b) successfully disentangles diverse semantic skills and autonomously synthesizes novel backward maneuvers.}
    \label{fig:umap_manifold}
\end{figure}

\begin{table}[!htbp]
\centering
\small
\caption{Ablation study on motion quality and safety.}
\label{tab:ablation}
\begin{tabular}{@{}lcc@{}}
\toprule
\textbf{Method Variant} & \textbf{FGD $\downarrow$} & \textbf{Safety Viol. $\downarrow$} \\
\midrule
Vanilla AMP                   & 4173.66 & Fail \\
\textbf{Diff-CAST (Ours)}     & 489.13  & \textbf{0} \\
\quad $\to$ w/o SACC          & \textbf{348.55} & 0 \\
\quad $\to$ w/o Safe Sim2Real & 1845.20 & 57 \\
\bottomrule
\end{tabular}
\end{table}

\FloatBarrier

As shown in Figure~\ref{fig:umap_manifold} and Table~\ref{tab:ablation},
Diff-CAST separates diverse semantic skills into well-structured latent
clusters and reduces FGD from 4173.66 to 489.13 compared with Vanilla AMP.
Although the w/o SACC variant obtains a lower FGD, the command-tracking results
below show that it sacrifices directional tracking robustness.

\paragraph{Omnidirectional Velocity Tracking}
Table~\ref{tab:command_tracking} evaluates tracking across forward, high-speed, lateral, and backward commands. Tracking error measures cumulative spatial deviation over task-specific distances (40\,m forward, 5\,m lateral/backward).

\begin{table}[htbp]
\centering
\small
\caption{Omnidirectional command tracking: position deviation and heading drift.}
\label{tab:command_tracking}
\begin{tabular}{lcccc}
\toprule
\multirow{2}{*}{\textbf{Target Command Profile}} & \multicolumn{2}{c}{\textbf{w/o SACC (Baseline)}} & \multicolumn{2}{c}{\textbf{Diff-CAST (Ours)}} \\
\cmidrule(lr){2-3} \cmidrule(lr){4-5}
& Pos. Dev. ($\downarrow$\,m) & Heading Drift ($\downarrow$\,rad) & Pos. Dev. ($\downarrow$\,m) & Heading Drift ($\downarrow$\,rad) \\
\midrule
Forward Walk ($v_x = 1.0$)         & 25.03       & 2.07  & \textbf{1.08}  & \textbf{0.13} \\
High-Speed Run ($v_x = 3.5$)       & 4.55        & 0.50  & \textbf{3.29}  & \textbf{0.38} \\
Pure Lateral ($v_y = 1.0$)          & 0.83        & 0.1   & \textbf{0.31}  & \textbf{0.04} \\
Pure Backward ($v_x = -1.0$)        & Fail (OOD)  & Fail (OOD) & \textbf{0.21}  & \textbf{0.16} \\
\bottomrule
\end{tabular}
\end{table}

\FloatBarrier

The unconditioned baseline fails under OOD commands, while Diff-CAST with SACC achieves near-zero heading drift across all directions. The w/o SACC variant achieves lower FGD by overfitting to forward-biased data, but inflates tracking error to 25.03\,m---confirming that SACC resolves the naturalness-vs-precision conflict.

\subsubsection{All-Terrain Locomotion}
\label{sec:exp_perceptive}
We evaluate the perceptive locomotion module on a procedurally generated
Level~0--9 terrain curriculum. The curriculum includes representative
unstructured terrains whose difficulty is progressively increased according to
the schedules in Table~\ref{tab:parkour_terrains}. Each test case is repeated
10 times, and Table~\ref{tab:terrain} reports the averaged success rate,
maximum level reached, bad impulse, unsafe foothold ratio, and energy.

\begin{table}[!htbp]
\centering
\small
\renewcommand{\arraystretch}{1.12}
\caption{Terrain curriculum for all-terrain locomotion evaluation.}
\label{tab:parkour_terrains}
\begin{tabular}{@{}p{0.1\textwidth}p{0.06\textwidth}p{0.66\textwidth}@{}}
\toprule
\textbf{Terrain} & \textbf{Levels} & \textbf{Difficulty schedule} \\
\midrule
Flat
& 0
& Flat baseline terrain without height noise or obstacles. \\

Rough Flat
& 0--9
& Random ground height perturbation increases from 0.01\,m to 0.08\,m. \\

Obstacles
& 0--9
& Low obstacle count increases from 8 to 60; obstacle size increases from 0.05\,m to 0.20\,m, with larger lateral offsets of 0.15--0.55\,m. \\

Slope
& 0--9
& Inclination angle increases from 10$^\circ$ to 35$^\circ$; both uphill and downhill directions are evaluated. \\

Stairs
& 0--9
& The staircase consists of 10 steps, with step height increasing from 0.10\,m to 0.30\,m and step length decreasing from 0.40\,m to 0.25\,m; both ascending and descending stairs are used. \\

Platform
& 0--9
& Single step-up or step-down height difference increases from 0.10\,m to 0.60\,m. \\

Gap
& 0--9
& Each terrain contains five gaps; gap width increases from 0.10\,m to 0.60\,m, with depth sampled from 0.50\,m to 0.80\,m. \\
\bottomrule
\end{tabular}
\end{table}

As shown in Table~\ref{tab:terrain}, Ours (full) achieves the best traversal performance, with an 83.2\% success rate and a Level score of 7.8.
Compared with the proprioception-only baseline, it reduces Bad Impulse from $90.0$ to 18.5~N$\cdot$s and Unsafe Foothold from $31\%$ to $14\%$, showing that the learned terrain-aware representation enables more deliberate and safer foothold selection.

\begin{table}[!tbp]
\centering
\caption{Quantitative ablation study on all-terrain traversal robustness, safety, and efficiency.}
\label{tab:terrain}
\resizebox{\textwidth}{!}{%
\begin{tabular}{lccccc}
\toprule
\textbf{Method} 
& \textbf{Success (\%)} $\uparrow$ 
& \textbf{Max Level} $\uparrow$ 
& \textbf{Bad Impulse (N$\cdot$s)} $\downarrow$ 
& \textbf{Unsafe Foothold (\%)} $\downarrow$ 
& \textbf{Energy$^{\dagger}$ (J)} $\downarrow$ \\
\midrule
Proprioception-only 
& 28.0 
& 2.2 
& 90.0 
& 31 
& 760 \\

\textbf{Ours (full)} 
& \textbf{83.2} 
& \textbf{7.8} 
& \textbf{18.5} 
& \textbf{14} 
& \textbf{895} \\

$\rightarrow$ w/o Memory 
& 72.4 
& 6.4 
& 30.0 
& 20 
& 930 \\

$\rightarrow$ w/o Ego-motion Compensation 
& 65.8 
& 5.6 
& 43.0 
& 18 
& 915 \\

$\rightarrow$ w/o Terrain Reconstruction Aux. 
& 68.6 
& 5.9 
& 38.0 
& 22 
& 925 \\
\bottomrule
\end{tabular}%
}
\vspace{2pt}
\begin{minipage}{0.98\textwidth}
\footnotesize
All values are averaged over the full terrain test set, with 10 repeated rollouts
per test case. $^{\dagger}$Energy depends on completed distance and terrain difficulty. Since the proprioception-only baseline often terminates early, energy is mainly compared among perceptive methods.
\end{minipage}
\end{table}

The ablations highlight the role of each component:
  \begin{enumerate}[label=(\alph*)]
    \item Removing memory lowers success to 72.4\% and increases Unsafe Foothold to 20\%, confirming that historical terrain context is critical for
  stable foothold selection.
    \item Removing ego-motion compensation causes the largest Bad Impulse increase (18.5 $\to$ 43.0~N$\cdot$s) due to spatial misalignment in historical
  LiDAR fusion.
    \item Removing terrain reconstruction losses reduces success to 68.6\% and raises Unsafe Foothold to 22\%, indicating that reconstruction
  supervision strengthens terrain-aware representations.
  \end{enumerate}
  Energy consumption is best interpreted alongside traversal outcome: the proprioception-only baseline consumes less energy primarily because it fails
  earlier. Among perceptive methods, Ours (Full) achieves the highest success and terrain level with the lowest energy.

\subsection{Scene Interaction}
\label{subsec:scene-interaction-experiments}

We evaluate scene interaction at two levels: IK reference generation and closed-loop execution after motion tracking. Table~\ref{tab:scene-interaction-stats} summarizes both stages. The IK statistics in Table~\ref{tab:scene-interaction-ik-stats} are used only to verify that the generated reaching references are geometrically feasible; the execution analysis in Table~\ref{tab:scene-interaction-error-comparison} is the main metric for assessing the deployed hand-shaking behavior.

\begin{table}[!tbp]
\centering
\small
\caption{Scene-interaction IK planning and execution error statistics.}
\label{tab:scene-interaction-stats}
\renewcommand{\arraystretch}{1.12}
\begin{subtable}[t]{0.52\textwidth}
\centering
\caption{IK reference planning accuracy. Error is the final Euclidean distance between the active front-foot site and the sampled hand target.}
\label{tab:scene-interaction-ik-stats}
\begin{tabular}{@{}p{0.5\linewidth}r@{}}
\toprule
\textbf{Quantity} & \textbf{Value} \\
\midrule
Reference segments & 2,048 \\
Mean / median error & 11.83\,mm / 9.83\,mm \\
90th percentile error & 24.55\,mm \\
Maximum error & 59.56\,mm \\
\bottomrule
\end{tabular}
\end{subtable}
\hfill
\begin{subtable}[t]{0.45\textwidth}
\centering
\caption{Error analysis for scene-interaction execution. Distances are millimetres; P90 denotes the 90th percentile.}
\label{tab:scene-interaction-error-comparison}
\setlength{\tabcolsep}{3.5pt}
\begin{tabular}{@{}lccc@{}}
\toprule
\textbf{Error} & \textbf{Mean} & \textbf{Median} & \textbf{P90} \\
\midrule
$e_{\mathrm{IK}}$ & 11.8 & 9.8 & 24.6 \\
$e_{\mathrm{track},C}$ & 52.9 & 34.4 & 125.5 \\
$e_{\mathrm{track},B}$ & 59.9 & 53.6 & 106.0 \\
$e_{\mathrm{exec}}$ & 137.4 & 121.5 & 193.1 \\
\bottomrule
\end{tabular}
\end{subtable}
\end{table}

For execution analysis, we separate local reaching performance into planning quality, tracking fidelity, and final endpoint accuracy. Let $p_G$ be the sampled hand target, $p_F^{\mathrm{IK}}$ the active-foot site in the generated IK reference, and $p_F$ the executed active-foot site. Let $p_C,p_C^{\mathrm{ref}}$ denote the executed and reference active front-calf body positions, and $p_B,p_B^{\mathrm{ref}}$ denote the executed and reference base-anchor positions. We report $e_{\mathrm{IK}}=\|p_F^{\mathrm{IK}}-p_G\|_2$, $e_{\mathrm{track},C}=\|p_C-p_C^{\mathrm{ref}}\|_2$, $e_{\mathrm{track},B}=\|p_B-p_B^{\mathrm{ref}}\|_2$, and $e_{\mathrm{exec}}=\|p_F-p_G\|_2$. These terms correspond to IK reference quality, local body tracking, base-anchor tracking, and the final hand-target placement error, respectively. Endpoint quantities are evaluated on hold frames, where the reference foot is intended to remain at the hand target.

The generated IK references are accurate at the centimetre scale, with a mean planning error of 1.18\,cm and a 90th percentile of 2.46\,cm. After policy execution, tracking remains stable but introduces additional local error: the active front-calf body has a 3.44\,cm median tracking error, and the base anchor has a 5.36\,cm median tracking error. The final executed active-foot-to-target error is 13.74\,cm on average and 12.15\,cm at the median, reflecting the combined effect of IK reference quality, dynamic tracking, and compliant whole-body execution during the hand-shaking hold phase.
\section{Applications}
\label{sec:applications}


Built on top of the unified control stack, an agent system serves as the high-level decision layer, operating in a full-duplex mode---continuously perceiving audio and video streams while simultaneously executing actions, enabling real-time interruption and adaptive response. It performs intent understanding and task decomposition, and dispatches the appropriate skill from the underlying behavioral layer. Task feedback is delivered through two complementary channels: physical body actions executed by the control stack, and synthesized speech responses.

In this section we demonstrate how the unified control stack presented in the preceding sections enables two representative downstream application scenarios: interactive companionship and intelligent navigation through complex environments.



\begin{figure}[htbp]
    \centering
    \includegraphics[width=\linewidth]{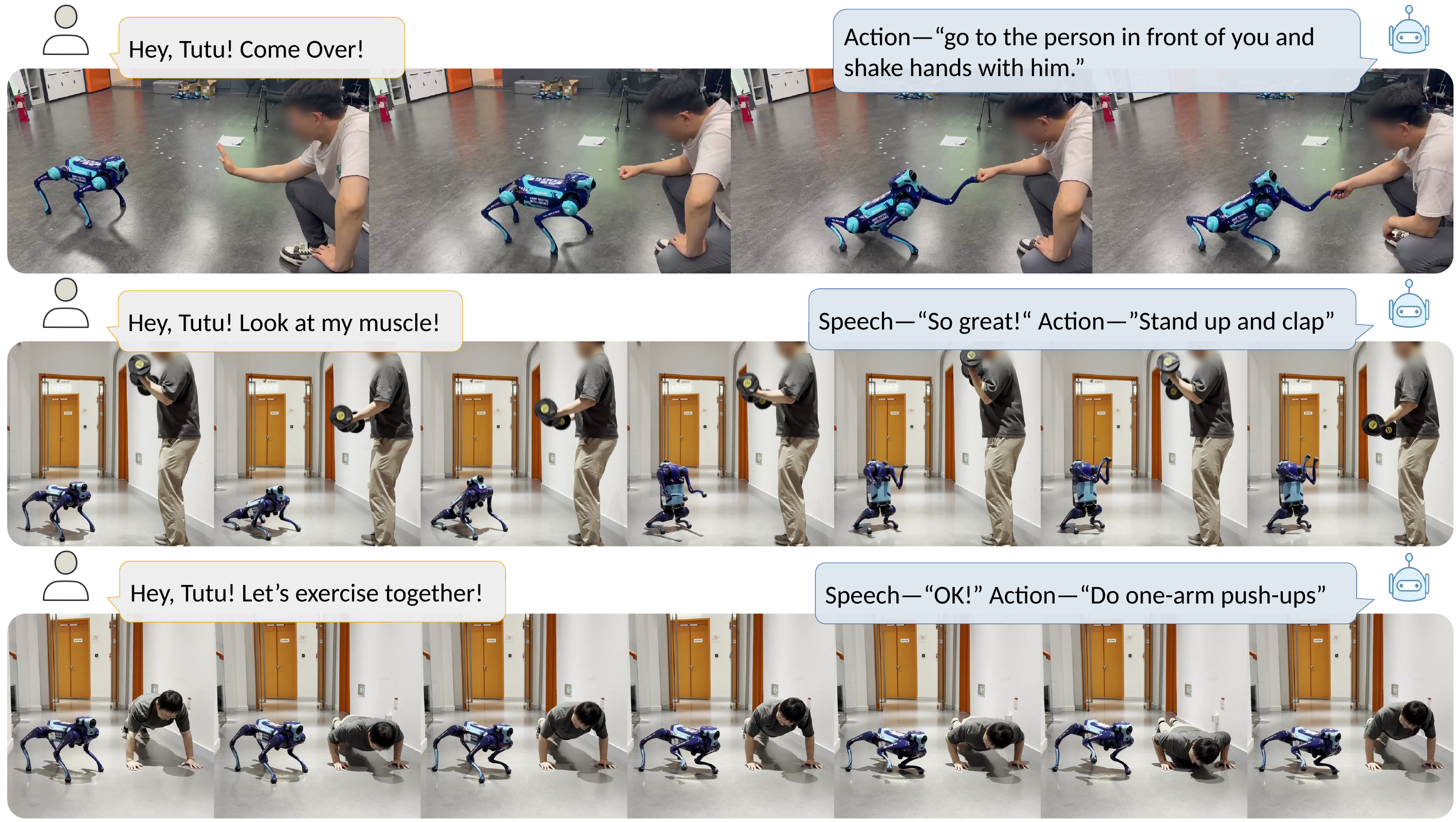}
    \caption{Examples of interactive companionship. Three representative scenarios are shown: (top) the robot approaches and shakes hands with the user upon request; (middle) the robot responds to a muscle-showing gesture with speech praise and a standing clap; (bottom) the robot joins the user in one-arm push-ups when invited to exercise together.}
    \label{fig:daily_case}
\end{figure}

\subsection{Interactive Companionship}

Interactive companionship requires the robot to interpret open-ended human intent and express the response through both speech and physically grounded body motion. Recent language-conditioned robotic systems and vision-language-action models demonstrate strong semantic reasoning and instruction-following capabilities with heavy ViT or VLM inference~\cite{ahn2022CanNotSay,driess2023PaLMEEmbodiedMultimodal,zitkovich2023RT2VisionlanguageactionModels,kim2024OpenVLAOpenSourceVisionLanguageAction,liu2026comatrack,gao2022DoublyfusedVitFuse}, but lightweight quadruped platforms cannot continuously host them together with real-time control, perception, and safety modules. ABot-C0 therefore separates cloud-side reasoning from on-robot control: computationally expensive multimodal understanding is offloaded to high-capacity cloud models, while motion execution, safety checks, and recovery remain within the onboard deployment stack. This design introduces communication and inference latency, but the latency is acceptable for low-frequency semantic reasoning and task planning, whereas high-frequency stabilization and policy execution stay local to the robot.
The resulting interaction pipeline works as follows:
\begin{enumerate}[label=(\arabic*)]
  \item The agent receives multimodal inputs, including audio and video streams, and offloads computationally heavy reasoning to cloud-side models for intent understanding, scene interpretation, and task decomposition, determining whether the user's request requires speech feedback or physical action.
  \item The cloud-side decision is translated into compact robot commands and dispatched to the on-robot control stack. Goal-directed instructions (e.g., ``come here'') are routed to the locomotion policy, while expressive action requests are converted to text commands and passed through a text-to-motion module that generates reference motion trajectories.
  \item The onboard unified control system executes the selected locomotion or motion-tracking policy in real time. Ongoing actions can be interrupted at any time, with the robot safely recovering to a stable standing state before accepting the next command.
\end{enumerate}
Representative daily interaction cases are shown in Fig.~\ref{fig:daily_case}, including approach-and-handshake behavior, gesture-aware response, and exercise companionship.

\subsection{All-Terrain Autonomous Navigation}

In navigation applications, the agent enters navigation mode where a VLN-based planner generates a sequence of waypoints, a local obstacle-avoidance module converts them into real-time velocity commands, and the control system executes accordingly. Meanwhile, perception and localization modules run continuously in the background, providing environment understanding and safety alerts.

\begin{figure}[!t]
    \centering
    \includegraphics[width=\linewidth]{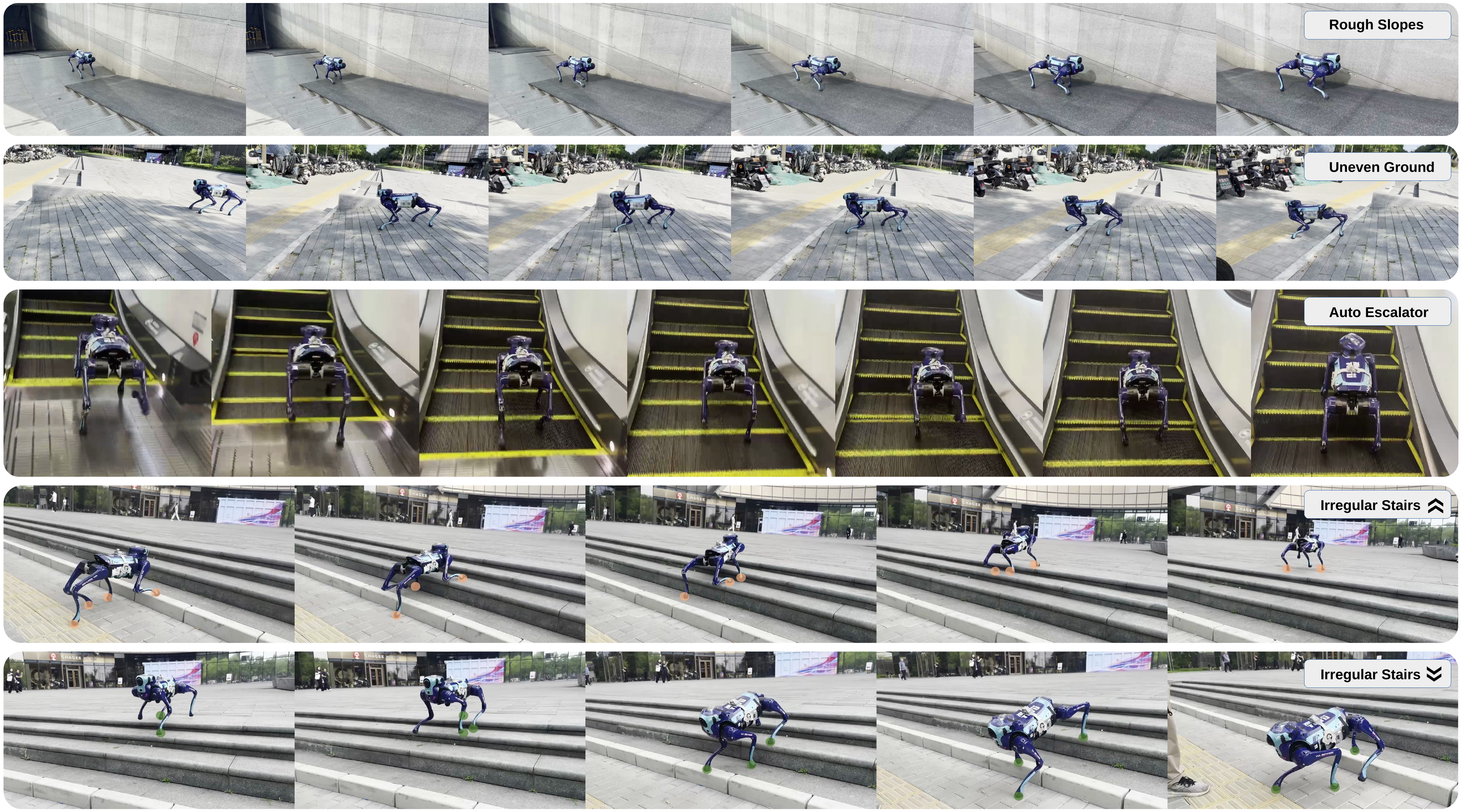}
    \caption{Demonstration of locomotion terrain adaptability. The top rows feature unstructured outdoor terrains (rough slopes and uneven ground). The middle row highlights dynamic indoor traversal on an auto escalator, while the bottom rows showcase complex step negotiation on irregular stairs (up) and irregular stairs (down).}
    \label{fig:loco_case}
\end{figure}

The control system receives velocity commands and adapts its gait in real time based on 3D point-cloud perception, enabling the robot to autonomously traverse complex outdoor environments including stairs, slopes, curbs, and uneven ground, which is shown in Fig.~\ref{fig:loco_case}.


\section{Conclusion, Limitations and Future Work}
\label{sec:conclusion}

We have presented \textbf{ABot-C0}, a system-level step toward behavior foundations for quadruped robots. ABot-C0 combines a scalable motion-data engine, generalist motion tracking, versatile locomotion, scene interaction, and a unified deployment stack. The data engine produces 16,074 physically validated motion trajectories, providing the scale and diversity needed to train a generalist motion-tracking policy with clear scaling behavior and strong zero-shot generalization. Together with robust locomotion and interaction policies, the deployment system supports real-world policy orchestration and downstream applications including interactive companionship and autonomous navigation.

ABot-C0 also exposes several limitations that define the next stage of quadruped behavior foundation models. First, the current system is still a coordinated multi-policy stack rather than a single unified model. Recent humanoid behavior foundation models and generalist robot policies suggest that future systems may benefit from unified architectures that condition on task intent, motion references, terrain observations, and interaction targets within one model family~\cite{bfmzero,holomotion,zitkovich2023RT2VisionlanguageactionModels,kim2024OpenVLAOpenSourceVisionLanguageAction,black2025pi0.5,sonic}. Second, ABot-C0 separates low-frequency semantic reasoning from high-frequency onboard control. This separation is practical for deployment, but the brain-body interface remains an open design problem: future systems should better handle command abstraction, latency, uncertainty, safety constraints, and recovery across this interface. Third, although ABot-C0 is validated on real hardware, broader generalization will require cross-platform evaluation, contact-rich interaction beyond the current case study, and stronger guarantees for hardware-safe deployment.

Looking forward, we view ABot-C0 as a foundation for a self-improving quadruped control stack. A promising direction is to close the loop between deployment and learning: robots should autonomously collect real-world experience, identify failures, expand their motion repertoire, and refine policies through continual skill discovery on hardware~\cite{zhao2026AgenticSkillDiscovery,xiao2026ENPIREAgenticRobot}. Combining scalable data generation, unified behavior modeling, and safety-aware real-world adaptation may ultimately turn quadruped robots from collections of isolated skills into extensible embodied agents.

\clearpage
\section{Contributions}
\label{sec:contributions}

Author contributions in the following areas are as follows:


\begin{multicols}{-2}
\subsubsection*{Writing}
    \begin{itemize}[itemsep=2pt, parsep=0pt, topsep=2pt, leftmargin=1.2em, label=\raisebox{0.1ex}{\tiny\textbullet}]
        \item Xufeng Zhao
        \item Fuzhi Yang
        \item Jianhui Chen
        \item Li Gao
    \end{itemize}

\subsubsection*{Data Pipeline}
    \begin{itemize}[itemsep=2pt, parsep=0pt, topsep=2pt, leftmargin=1.2em, label=\raisebox{0.1ex}{\tiny\textbullet}]
        \item Li Gao
        \item Jianhui Chen
        \item Yao Zheng
    \end{itemize}

\subsubsection*{Model Training and Evaluation}
    \begin{itemize}[itemsep=2pt, parsep=0pt, topsep=2pt, leftmargin=1.2em, label=\raisebox{0.1ex}{\tiny\textbullet}]
        \item Fuzhi Yang
        \item Jianhui Chen
        \item Xufeng Zhao
        \item Zhang Meng
        \item Jie Gao
        \item Wenyu Liu
        \item Congyang Zhao
        \item Tianxiong Lv
    \end{itemize}

\subsubsection*{Algorithm Architecture}
    \begin{itemize}[itemsep=2pt, parsep=0pt, topsep=2pt, leftmargin=1.2em, label=\raisebox{0.1ex}{\tiny\textbullet}]
        \item Xufeng Zhao
        \item Fuzhi Yang
        \item Jianhui Chen
    \end{itemize}

    \columnbreak

\subsubsection*{Deployment}
    \begin{itemize}[itemsep=2pt, parsep=0pt, topsep=2pt, leftmargin=1.2em, label=\raisebox{0.1ex}{\tiny\textbullet}]
        \item Jianhui Chen
        \item Xufeng Zhao
        \item Fuzhi Yang
    \end{itemize}

\subsubsection*{Hardware \& System}
    \begin{itemize}[itemsep=2pt, parsep=0pt, topsep=2pt, leftmargin=1.2em, label=\raisebox{0.1ex}{\tiny\textbullet}]
        \item Menglin Yang
        \item Minqi Gu
        \item Yaru Zhao
        \item Honglin Han
    \end{itemize}

\subsubsection*{Interaction Experience}
    \begin{itemize}[itemsep=2pt, parsep=0pt, topsep=2pt, leftmargin=1.2em, label=\raisebox{0.1ex}{\tiny\textbullet}]
        \item Shihui Su
        \item Zixiao Tang
    \end{itemize}

\subsubsection*{Project Leads}
    \begin{itemize}[itemsep=2pt, parsep=0pt, topsep=2pt, leftmargin=1.2em, label=\raisebox{0.1ex}{\tiny\textbullet}]
        \item Liu Liu
        \item Yang Cai
    \end{itemize}

\subsubsection*{Advisors}
    \begin{itemize}[itemsep=2pt, parsep=0pt, topsep=2pt, leftmargin=1.2em, label=\raisebox{0.1ex}{\tiny\textbullet}]
        \item Yang Cai
        \item Wenbin Tang
        \item Mu Xu
    \end{itemize}
\end{multicols}

\textbf{Acknowledgement:}  We thank all contributors to the Tutu project for their support throughout this work. We also thank Yifei Qian for support.



\clearpage

\bibliographystyle{plainnat}
\bibliography{ref,report}

%

\end{document}